\documentclass[prodmode,acmjocch]{acmlarge}

\usepackage{amsmath}
\usepackage{mathtools}
\usepackage{amssymb}

\usepackage{graphicx}
\usepackage[caption=false]{subfig}
\usepackage{hyperref}
\usepackage{breakurl}	

\acmVolume{2}
\acmNumber{3}
\acmArticle{1}
\articleSeq{1}
\acmYear{2016}
\acmMonth{2}

\usepackage[ruled]{algorithm2e}
\SetAlFnt{\algofont}
\SetAlCapFnt{\algofont}
\SetAlCapNameFnt{\algofont}
\SetAlCapHSkip{0pt}
\IncMargin{-\parindent}
\renewcommand{\algorithmcfname}{ALGORITHM}

\title{God(s) Know(s): Developmental and Cross-Cultural Patterns in Children Drawings}
\author{KSENIA KONYUSHKOVA, NIKOLAOS ARVANITOPOULOS and SABINE S\"{U}SSTRUNK \affil{School of Computer and Communication Sciences, \'{E}cole Polytechnique F\'{e}d\'{e}rale de Lausanne (EPFL), Switzerland}
ZHARGALMA DANDAROVA ROBERT and PIERRE-YVES BRANDT \affil{Institute for Social Sciences of Contemporary Religion, University of Lausanne (UNIL), Switzerland}
}

\begin{abstract}
This paper introduces a novel approach to data analysis designed for the needs of specialists in psychology of religion. 
We detect developmental and cross-cultural patterns in children's drawings of God(s) and other supernatural agents. 
We develop methods to objectively evaluate our empirical observations of the drawings with respect to: (1) the gravity center, 
(2) the average intensities of the colors \emph{green} and \emph{yellow}, (3) the use of different colors (palette) and (4) the visual 
complexity of the drawings. We find statistically significant differences across ages and countries in the gravity centers and in the average 
intensities of colors. These findings support the hypotheses of the experts and raise new questions for further investigation.
\end{abstract}



\acmformat{Ksenia Konyushkova, Nikolaos Arvanitopoulos, Zhargalma Dandarova Robert, Pierre-Yves Brandt and Sabine S\"{u}sstrunk. 2014. God(s) know(s): developmental and cross-cultural patterns in children drawings.}

\begin{document}

\graphicspath{{./pics/}}

\begin{bottomstuff}
This work is supported by 
\end{bottomstuff}

\maketitle

\section{Introduction}
\label{sec:intro}
\begin{figure}
  \centering
  \subfloat[Drawing from Russia.]{\includegraphics[width=0.45\columnwidth]{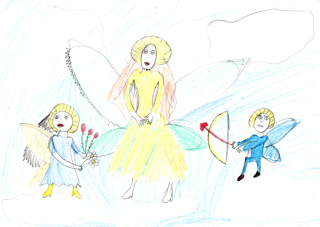}} \hspace{0.5in}
  \subfloat[Drawing from Japan.]{\includegraphics[width=0.45\columnwidth]{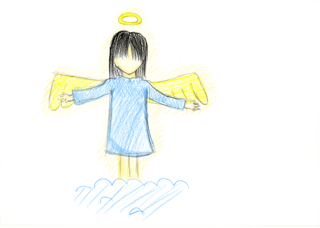}}
  \caption{Two examples from the dataset of children's drawings of God(s).}
  \label{fig:img_dataset}
\end{figure}

This article presents our computer-aided approach to a psychology-oriented analysis of children's drawings of God(s). 
The purpose of this study is to detect developmental and cross-cultural patterns in bitmapped data. 
Given a dataset of digital scans of children's drawings of God(s), our task is to search for appropriate imaging tools and to 
apply them to (1) find evidence for or against hypotheses from the psychology of religion and cognitive developmental psychology and to (2) provide insights to formulate new hypotheses. 
The unique drawing dataset was collected under the direction of Pierre-Yves Brandt \cite{brandt:10}, \cite{brandt:09}, \cite{dandarova:13} and \cite{ladd:98}. 
Figure~\ref{fig:img_dataset} shows two examples of drawings from the dataset.

Drawing analysis is a well-studied field in psychological studies with children: it is used in cognitive and clinical assessment and personality research. 
Several developmental studies in child psychology based on drawings are proposed in the literature \cite{strommen:95}, \cite{reiss:01}, \cite{kose:08}, \cite{oskarsdottir:11}. 
Many authors believe that the activity of drawing plays an important role in the child's cognitive development. 
Antonio Mach{\'o}n \cite{machon:13} provides a review on the development of graphic representation by pre-school and primary-school children ($1-10$ years old). 

Our research in psychology of religion aims to understand, through children's drawings, the origin and development of the concept of God(s). 
The first study by Ernest Harms dates back to $1944$, when he collected more than $4800$ drawings of God(s) by children from $3-16$ years \cite{harms:44}. 
He distinguished three stages of progression in the children's religious development and demonstrated the effect of children's age on their notion of God(s). 
Other studies on age patterns include research on anthropomorphous versus symbolic God(s) representations  
\cite{pitts:77}, \cite{eshleman:99}, \cite{hanisch:96}, \cite{tamm:96}, 
\cite{ladd:98}, \cite{brandt:09}, \cite{dandarova:13}.
Hanisch studies the influence of cultural and religious contexts on drawings of children in Western and Eastern Germany \cite{hanisch:96}.
A few authors dedicate their research to comparisons between drawings from different Christian groups -- Mormon, Lutheran, 
Catholic, Southern Baptist, Pentecostal and Roman Catholic \cite{pitts:77}, \cite{ladd:98}.
Even though Western Christian cultures are the most popular subjects, several works exist about other cultural and 
religious contexts, such as those of Japan \cite{brandt:09} and Eastern Siberia \cite{dandarova:13}. 
Differences in representations of God(s) between male and female participants are also noted in some gender studies  
\cite{hanisch:96}, \cite{ladd:98}, \cite{kay:04}, \cite{brandt:09}. There also exist studies of adults' drawings of God(s) 
\cite{newberg:10}, \cite{rizzuto:81}, \cite{goodman:08}

The issue of children's representations of supernatural agents has thus become a popular subject of research in the psychology of religion. 
Compared to quantitative methods (such as questionnaires), qualitative measures (open-ended tasks, such as drawings) enable us to investigate 
complicated and individual notions of God(s). However, with more flexibility, new difficulties for analyzing the data emerge. 
The first problem is the amount of data that requires manual processing by experts. 
For example, to obtain statistics on the proportion of colored and gray-scale drawings by Buryat children, the experts had to manually access all the images \cite{dandarova:13}. 
A more high-level problem that is not easy to detect arises due to the cultural sensitivity of those who conduct the research. Being a part of a specific culture, the experts 
design descriptors that are biased for analysis: these descriptors are trends for a particular culture, religion and epoch \cite{hanisch:96}.
These types of culture-centric descriptors are not well-suited for inter-cultural research and there is need for features that are culturally independent. 

From all the above studies, it is clear that the researchers lack automated tools for large-scale developmental and cross-cultural studies. 
In this paper, we aim to develop computational methods for formalizing and testing our hypotheses about children's cognitive development and environmental influence. 
The growth of the dataset from $N=142$, at the beginning of the project, to $N=2389$ drawings necessitates the use of automatic tools 
in visual analysis \cite{brandt:10}, \cite{brandt:09}, \cite{dandarova:13}, \cite{ladd:98} as manual evaluations over this number of drawings become infeasible.  
The task we want to address is quite distinct from the usual tasks in computer vision, due to the content of the dataset: images of drawings have very 
few similarities with images of natural scenes. Drawings have a more discrete nature than natural images, which makes common techniques used in computer vision 
fail. For example, a common method for natural images, such as edge detection, detects stroke edges rather than object edges. 
Another example is the difference in color gamut: natural images have a much 
broader color range with fine quantization compared to drawings. Moreover, the absence of clear boundaries between objects renders the application of 
computer vision algorithms quite challenging.

Computer vision, image processing, and data mining can be a source of automated methods for graphical studies in the psychology of religion, 
but the methods that can be applied to drawings are not numerous. 
Usually, the algorithms were developed for photographic or scientific imaging, hence the methods for drawing analysis are quite limited. 
In his work, Stork \cite{stork:09} provides a review of the computer vision, image processing and computer graphics applications to the analysis of art works (paintings, frescos, etc.), 
which includes brush strokes analysis, dewarping, image enhancement, lighting and illumination analysis, and perspective and composition analysis. 
However, the analysis of developmental and cultural factors through children's drawings has not been addressed before. Existing methods for the analysis of paintings 
are not necessarily applicable to a dataset of children's drawings, because (1) their content and form substantially differs from those of paintings and (2) the above mentioned 
applications are not related to the research task we address in this work. 

In the following sections, we describe analytical methods for pattern analysis that we apply to the acquired dataset of children's drawings. 
We call this analysis \emph{exploratory}, because even though we have some hypotheses about our data, we do not have a specific pattern to look for. 
Our hypotheses come from our intuitive observations of the drawings. Our goal is to test them with image analysis tools and see if they receive support through objective methods. 
We investigate four different observations: (1) the location of the gravity center, (2) the average intensities of the colors \emph{green} and {yellow}, 
(3) the use of different colors inside the drawings, and (4) the visual complexity of the drawings.
These investigations are based on general questions of researchers who study the religious psychology of children through their drawings, such as 
the location of the drawing on the sheet, the type and number of colors used, the complexity of the drawing, the drawing differences across cultures and ages, etc.
We specifically aim to link the empirical evidence that we derive by visual inspection of the drawings with objective measures.

Our paper is organized as follows. In Section \ref{sec:data}, we describe the dataset to which we apply our automatic analysis and the data collection process. 
In Section \ref{sec:datamining} we discuss our hypotheses and the automatic methods we used for pattern extraction and analysis on the drawing dataset. 
Finally, in Section \ref{sec:concl} we conclude our work with discussions and future work directions.

\section{Dataset} 
\label{sec:data}
The dataset\footnote{\burl{http://www.unil.ch/issrc/home/menuinst/recherches/psychologie-de-la-religion/developpent-religieux-et-spi/english-version.html}} 
consists of $2389$ children's drawings of God(s), collected by Pierre-Yves Brandt, Zhargalma Dandarova and their students \cite{brandt:10}, \cite{dandarova:13}, 
and additional drawings collected by \cite{ladd:98}. It is a unique digital source of children's drawings of God(s). 

Every child was given the following material: a blank A4 sheet, a box of water-resistant wax crayons 
($8$ to $12$ colors: blue, green, red, orange, yellow, brown, black, white; additional colors: purple, grey, light blue, dark blue, light green and dark green). 
If no wax crayons were available, color pencils (of the same colors as above) and erasers were provided.

The children were asked to write their name on one side of the sheet and to draw on the other side. The instruction of the task was given orally: 
``Have you ever heard of the word ``God''? Try to imagine and draw it on your sheet''. 
Furthermore, they had to answer a questionnaire about their religious knowledge and to explain how they performed the task. 
A small group of children in Switzerland was asked to perform a related, but a slightly different and 
more general task: to draw what they think of when they think about God(s). 
In our analysis, we denote the former part of a dataset as ``God(s)" and the latter part as ``General". 
Also, the children from the US received a slightly different task: to ``draw a picture of God''.
The scans contain meta-data information about the country and the region in which the drawings were collected, the school (religious or not religious), 
as well as the age and gender of children.
\begin{figure}[h]
  \centering
  \subfloat[Number of drawings per country: Switzerland (CH), Japan (JP), Romania (RO), Russia (RU), USA (US).]{\includegraphics[width=0.42\columnwidth]{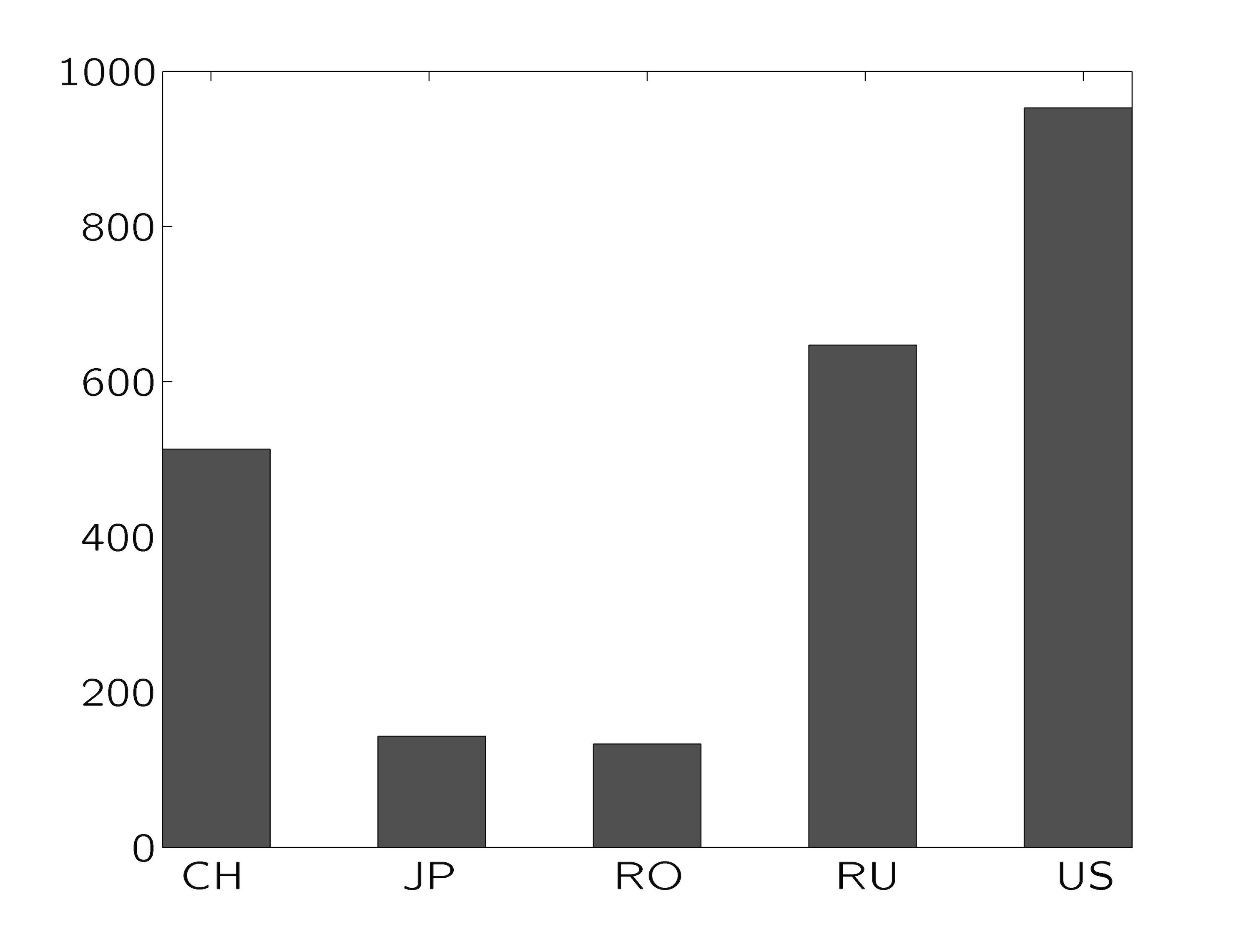}  \label{pic:countries}}\hspace{0.5in}
  \subfloat[Distribution of children's ages.]{\includegraphics[width=0.42\columnwidth]{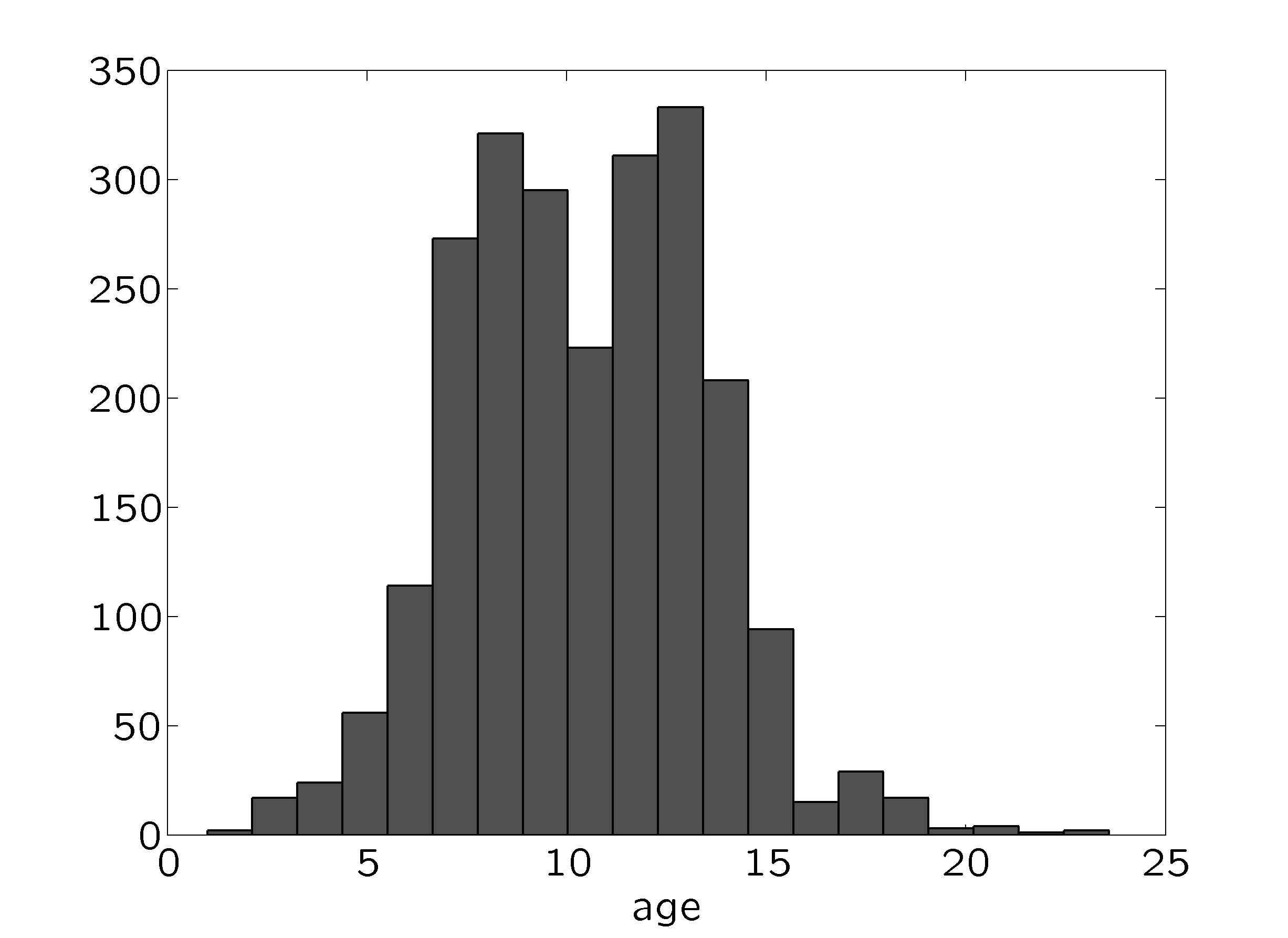}\label{pic:ages}}
  \caption{Distribution of children's drawings in the dataset according to countries and age groups.} 
\end{figure}

The drawings were obtained from multiple geographical locations: Switzerland (CH), Japan (JP), Romania (RO), Russia (RU), and the USA (US). 
The religions of these countries include Orthodoxy, Catholicism, Protestantism, and Buddhism. Figure \ref{pic:countries} illustrates the distribution of the 
number of drawings from different countries. The dataset is not balanced: there are many images from Switzerland, Russia and USA, but only a few from Japan and Romania.  
Let us mention here that in our analysis, we sometimes explicitly split the Russian drawings into two subsets corresponding to two regions: Buryatia (RU-bo) and Saint-Petersburg (RU-sp). 
The regions are quite far from each other and religious traditions are different enough to expect significant differences in the data.

Most of the children who participated in the experiments were from $5$ to $15$ years old (see Figure \ref{pic:ages}). 
We split the dataset into the following age categories: $(1,7]$, $(7,9]$, $(9,11]$, $(11,13]$ and $(13,23]$. 
Such a division ensured that the subsets were of approximately the same size and that they contained a significant number of drawings on which statistical testing 
is meaningful. More information about the dataset can be found on the project's website.

\section{Automatic Pattern Analysis} 
\label{sec:datamining}

\subsection{Gravity Analysis}
\label{sec:gravityanalysis}
The usual gravity center seen in standard children's drawings is in the center. \cite{winner:06} reports that ``the centering principle can be seen in drawings of 3-year-olds. 
The earliest use of centering consists of a single figure placed in the middle of the page.'' This results in symmetry, which increases with age. 
Thus, we are interested in studying if and how religious ideas and representations influence the position of the gravity center in the drawings. 
We hypothesize that the gravity center of drawings of God(s) is shifted towards the upper part of the image, i.e., children draw mostly on the upper part of the paper 
and not on the lower part or on the sides. The intuition behind this hypothesis is that most children believe that God(s) lives
in the sky, therefore they have a tendency to draw their notion of God(s) slightly above the center of the page. In this section, 
we thus investigate whether in drawings of God(s) the gravity center is located above the middle of the page. 

To quantify this hypothesis, we use the following procedure to determine the gravity center. First, we convert the images to gray scale by taking average values of the RGB channels. 
Then, we scale each image $i=1,...,N$ to be of height $h$. The resulting image is
\[
\mathbf{M}^{(i)} = 
\begin{pmatrix}
  m_{1,1}^{(i)} & m_{1,2}^{(i)} & \cdots & m_{1,w_i}^{(i)} \\
  m_{2,1}^{(i)} & m_{2,2}^{(i)} & \cdots & m_{2,w_i}^{(i)} \\
  \vdots  & \vdots  & \ddots & \vdots  \\
  m_{h,1}^{(i)} & m_{h,2}^{(i)} & \cdots & m_{h,w_i}^{(i)}
\end{pmatrix}.
\]
We compute the vector $\mathbf{g}^{(i)}\in \mathbb{R}^h$, which contains the average intensity of pixels in each of the $h$ rows of $\mathbf{M}^{(i)}$, as follows:
\[
g_j^{(i)} = 255 - \frac{1}{w_i} \sum_{k=1}^{w_i} m_{jk}^{(i)}, \quad j = 1,\dots,h.
\]
The vector $\mathbf{g}^{(i)}$ for image $\mathbf{M}^{(i)}$ has values in the interval $[0,255]$, where lower values correspond to the positions with few colored pixels 
and larger values correspond to those positions with more colored pixels in it. 

Finally, to analyze statistical patterns in the data, we partition the dataset into meaningful groups $P_1, P_2, ... , P_k$, where $k$ depends on the context of a task, 
such as the number of age groups or the number of geographical regions. Each image belongs to some group $P_s$, $s=1,...,k$. 
For $s = 1,...,k$ we consider sets of images that belong to each group $\mathbf{G}^{(s)} = \{\mathbf{g}^{(i)} \mid i \in P_s \}$. We compute the average intensity vectors  
$\mathbf{\overline G}_s$ per group as follows:
\[
 \mathbf{\overline G}_s = \frac{1}{\mid P_s \mid} \sum_{i \in P_s} \mathbf{g}^{(i)}.
\]

\subsubsection{Gravity analysis for data segments}
\label{sec:gravitydatasegments}
\begin{figure}[h]
  \centering
  \subfloat[CH]{\includegraphics[width=0.3\columnwidth]{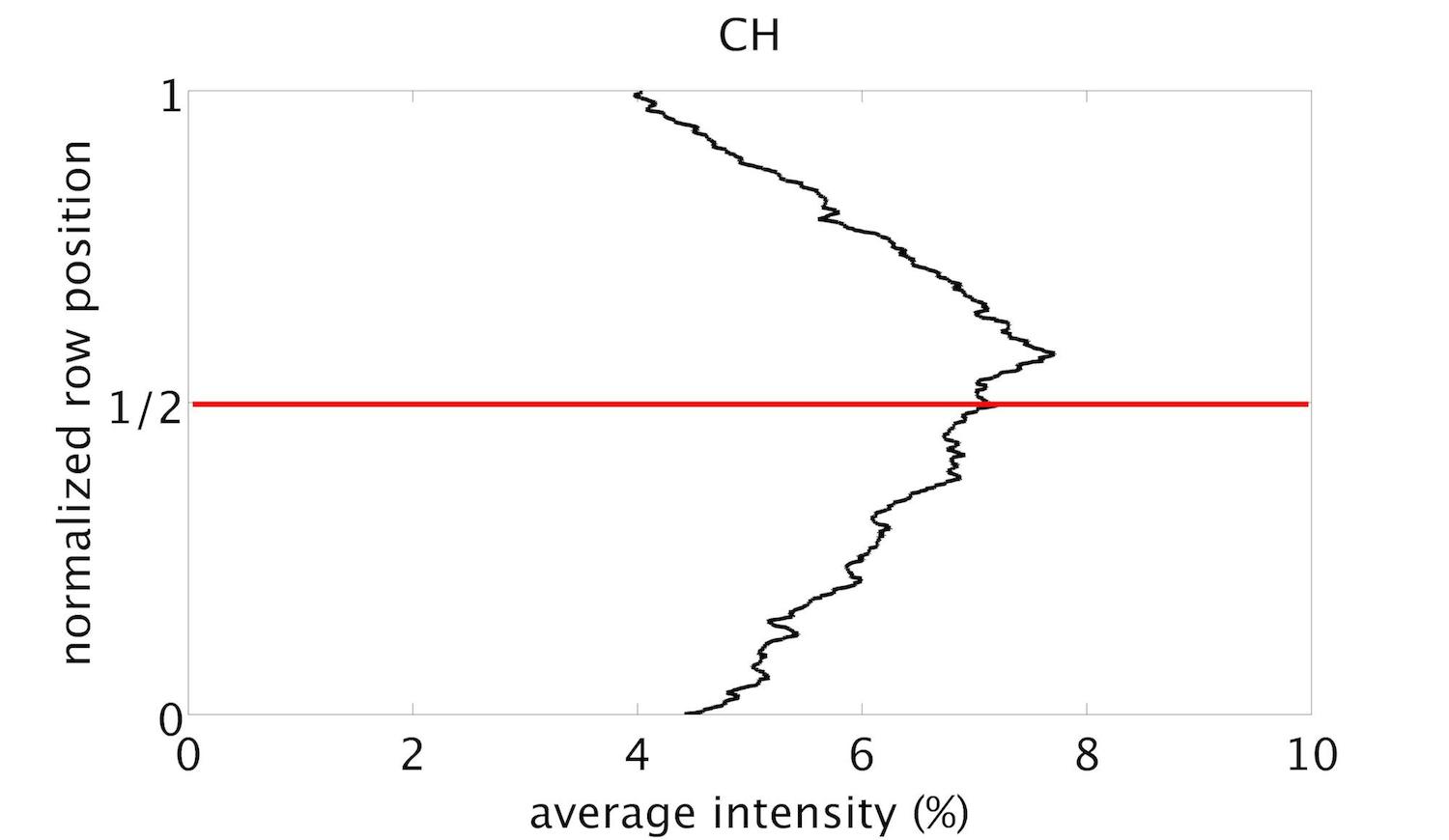}}
  \subfloat[JP]{\includegraphics[width=0.3\columnwidth]{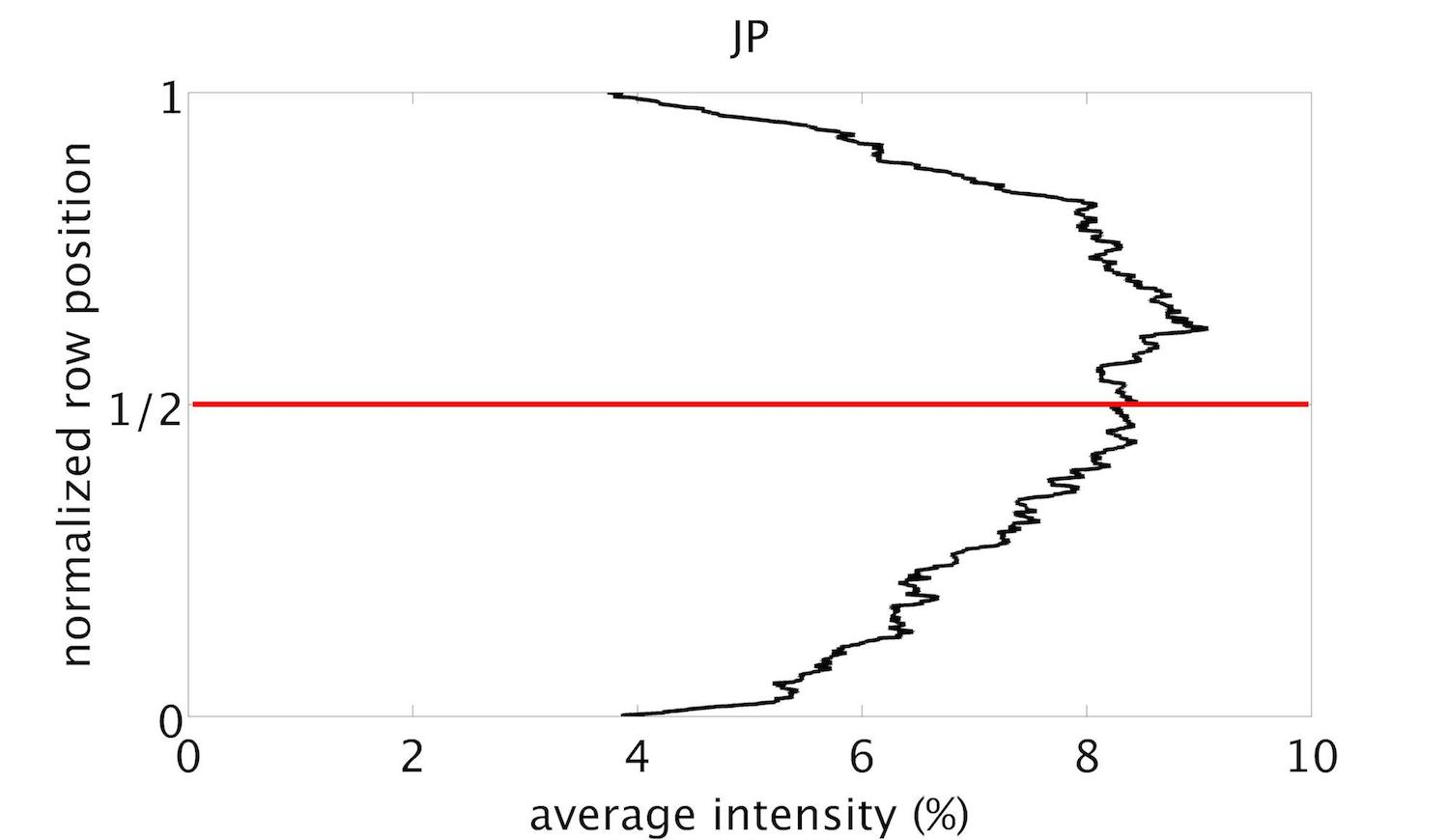}}
  \subfloat[RO]{\includegraphics[width=0.3\columnwidth]{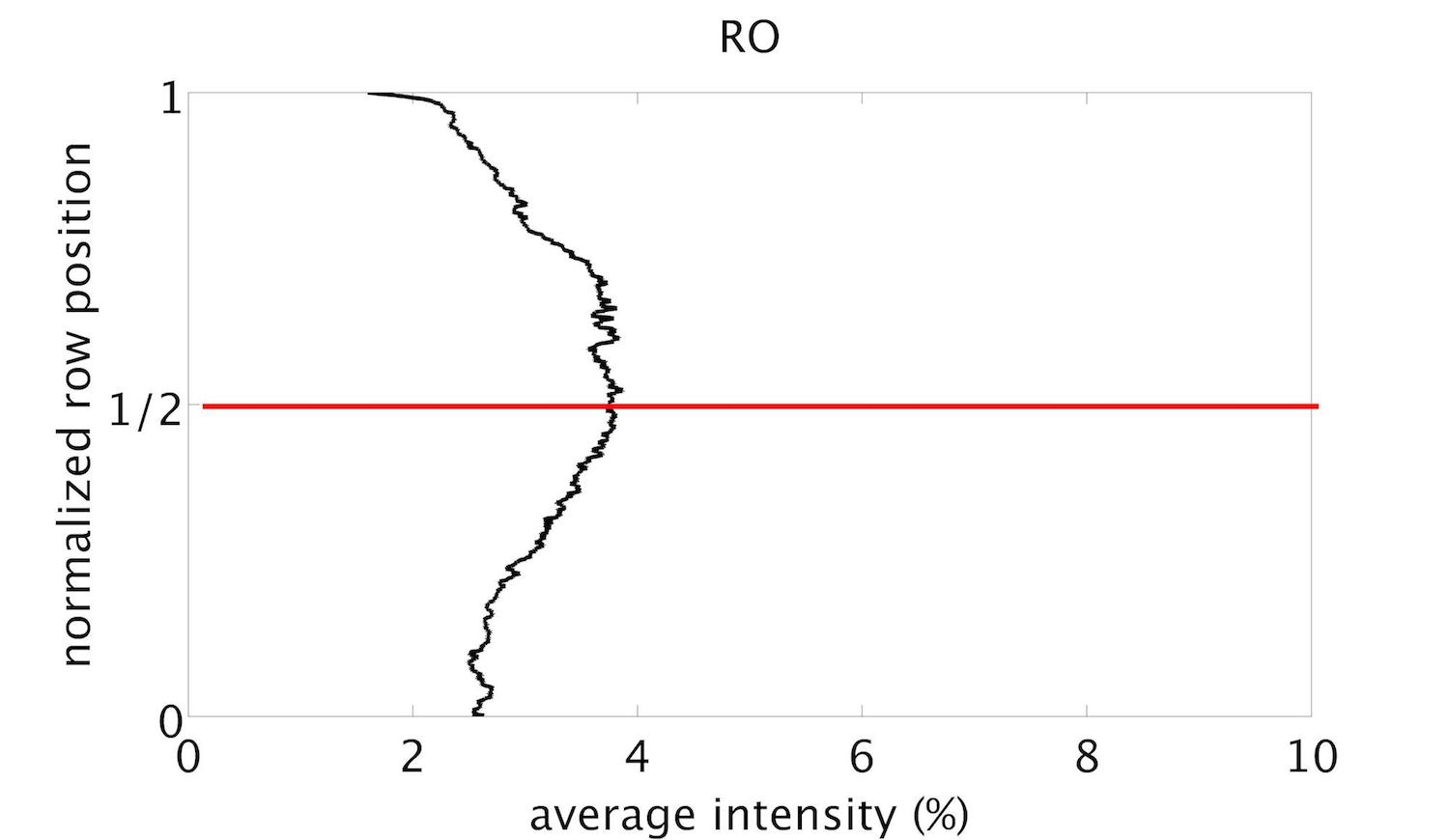}} \\ \vspace{0.1in}
  \subfloat[RU-bo]{\includegraphics[width=0.3\columnwidth]{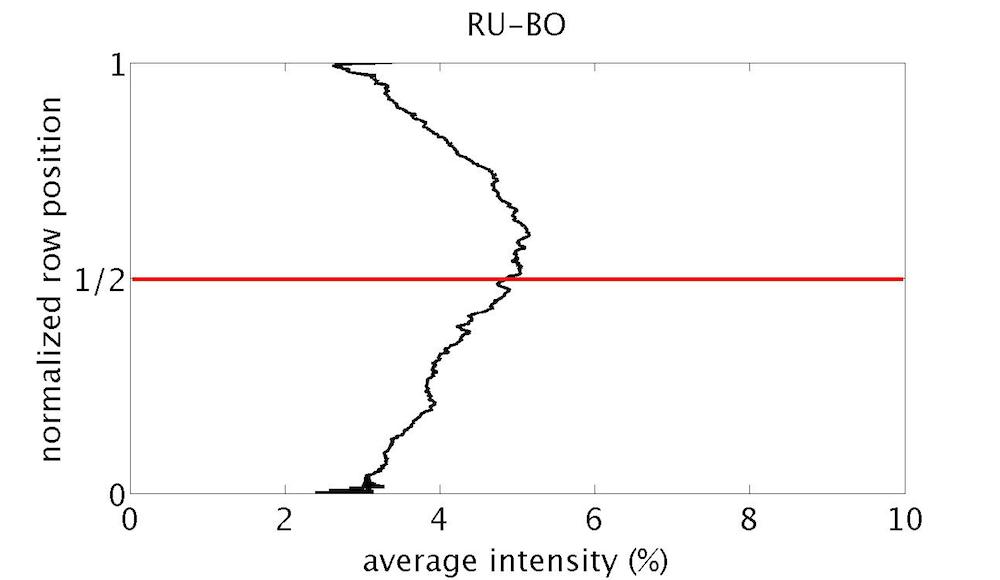}}
  \subfloat[RU-sp]{\includegraphics[width=0.3\columnwidth]{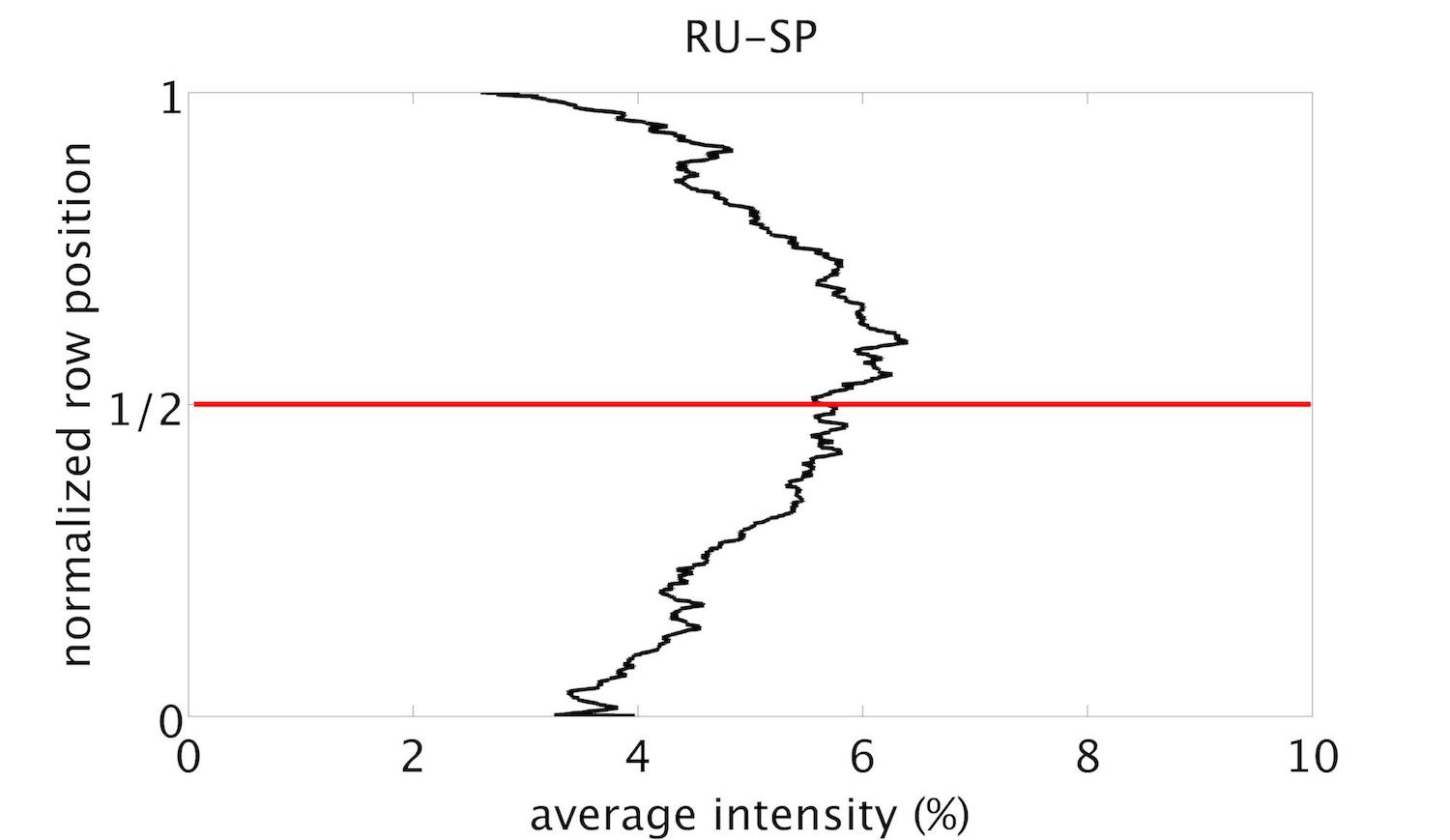}}
  \subfloat[US]{\includegraphics[width=0.3\columnwidth]{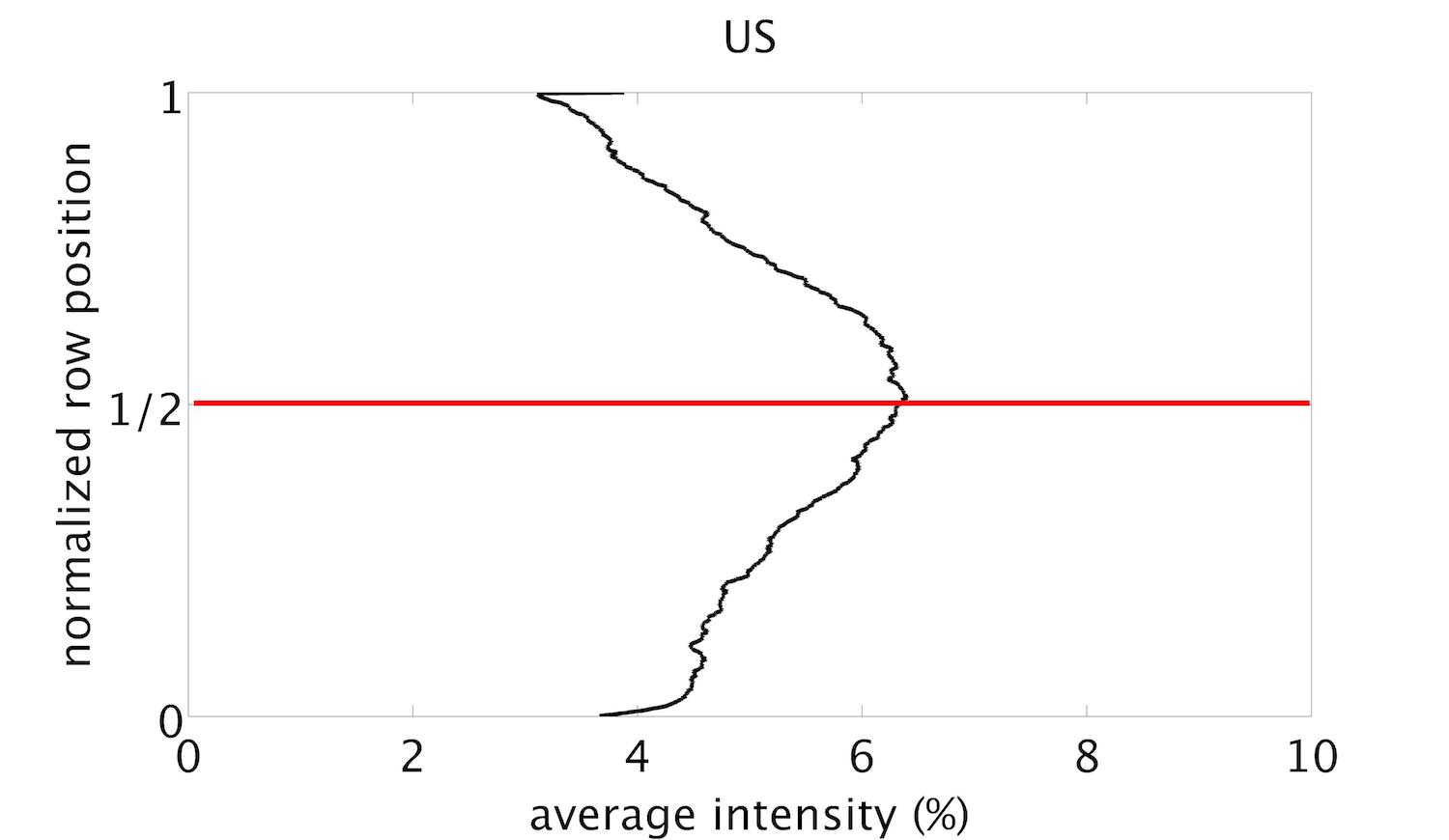}}
  \caption{Average intensity vectors in different countries. We observe that the gravity center is often biased to the upper part of the drawings.} 
  \label{pic:gravity-countries}
\end{figure}

We compare the average intensity vectors between regions and between ages. 
First, we test the cultural influence imposed by different regions and religious traditions. We split the dataset into country groups as discussed in 
Section \ref{sec:data} into Switzerland (CH), Romania (RO), Russia Buryatia (RU-BO), Russia Saint-Petersburg (RU-SP) and the USA (US). 
We calculate the average intensity vectors for these countries and visualize them in Figure \ref{pic:gravity-countries}. 
The horizontal axis shows the percentage of the average inverse intensity of pixels for each row of the image matrix, and the vertical axis denotes the normalized row position.

Empirically, we observe that the shapes of the average intensity vectors look different. For instance, in the drawings from Switzerland, Japan and Russia, 
the gravity center tends to be higher than the center of the image. This evidence partially verifies the hypothesis about the shifted gravity center  
in children's drawings of God(s). It provides evidence that children express their idea that the concept of God(s) is associated 
with what is high, up, above, on the top. On the contrary, the gravity center of the US drawings is in the middle. 
It seems that, for these children, the composition strategy to draw God(s) does not differ from the strategy to draw anything else. This may be an indication that the concept of God(s) 
is considered a common concept for the children in this sample (US children, from eight Christian groups, attending religious education classes in the late nineties). 
It could also be due to the fact that these children received slightly different instructions. In fact, they were only asked to ``draw a picture of God'' \cite{ladd:98}. 
Further investigation is necessary to clarify this result. The same goes for other cross-cultural variations that can be observed: for example, 
the difference in the intensity proportion between the least colored parts of the drawings (top and bottom) and the more colored parts (mid or above-mid of the drawings). 

It is still too early to conclude that cultural traditions have such an impact on the difference in the above proportions. 
We can think of several possible interpretations for this difference: the diffuse presence of God(s) in some drawings, 
the possible pictorial traditions of coloring, or only sketching, a figure, the importance of the color white in some cultures, etc. 
This example demonstrates that our automatic approach in this type of research not only tests existing questions and hypotheses, but also serves as a source for generating new ones.

\subsubsection{Average intensity differences by country}
\label{sec:gravitycountry}
\begin{figure}
  \centering
  \subfloat[Statistical test results for geographical regions.]{\includegraphics[width=0.4\columnwidth]{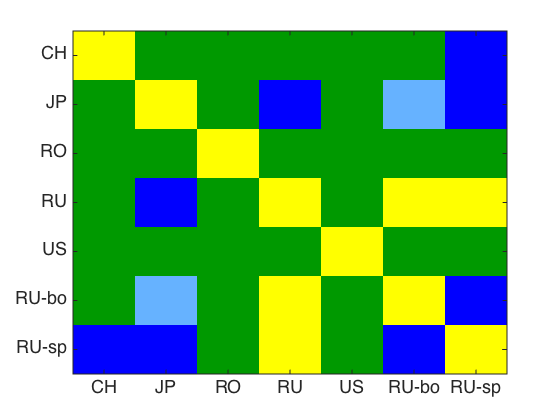}\label{pic:stats-geo}}\hspace{0.5in}
  \subfloat[Statistical test results for age groups.]{\includegraphics[width=0.4\columnwidth]{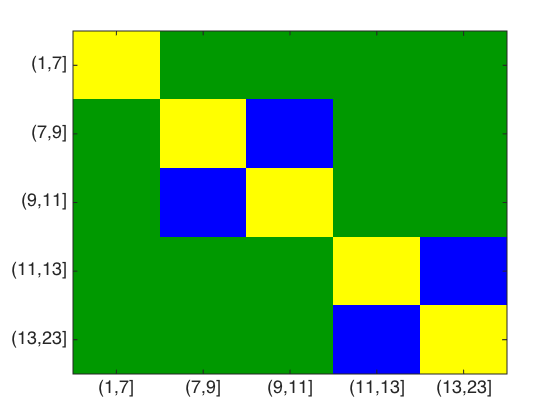}\label{pic:stats-age}}
  \caption{Statistical test results ($p$ values) for the average intensity vectors. Green denotes the value $0$ and light blue denotes values in the range $[0,0.05]$ (significant differences at $95\%$ confidence level). Blue denotes values greater than $0.05$ (not significant differences at $95\%$ confidence level.)}
  \label{pic:stats} 
\end{figure}

Having obtained the results that give evidence to some pattern in the average intensity vectors, it is important to verify that it is unlikely that this occurred by chance alone.
To make conclusions about the psychological or theological findings in the data, we need to test the significance of the results with statistical-significance tests. 
That is, we want to test if the groups $\mathbf{G}^{(s)}$ of average intensity vectors are sampled from the same distribution. 
Our data is non-Gaussian, therefore, we need a non-parametric test for a multivariate distribution. 
The chosen method is a multivariate two-sample test, which is a type of permutation test based on the number of coincidences of nearest neighbours \cite{henze:88}. 
It considers two independent $d$-dimensional samples from distributions $f_1$ and $f_2$ and tests the hypothesis $f_1 = f_2$. 
The test does not depend on the form of the distributions $f_1$ and $f_2$, as the number of nearest-neighbour coincidences has a limiting normal distribution.
We set the $p$-value to $0.05$, that is, we reject the hypothesis of two distributions being equal at $95\%$ confidence level.
Figure \ref{pic:stats-geo} shows the statistical test results between all pairs of countries. 
The differences are significant at $95\%$ confidence level for all country pairs, except Saint-Petersburg 
and Japan ({\it p-value} is $18.45\%$), Saint-Petersburg and Switzerland ($7.08\%$). Saint-Petersburg and Japan ($35.4\%$) and the two regions in 
Russia -- Saint-Petersburg and Buryatia ($17.98\%$). Note, however, that we cannot be completely sure that the above mentioned distributions are different. Statistical 
uniformity could be a result of the limited data and we might be able to prove the difference once the dataset is extended with more drawings.

\subsubsection{Average intensity differences by age}
\label{sec:gravityage}
Additionally, we examine the average intensity vectors between different age groups ($(1,7]$, $(7,9]$, $(9,11]$, $(11,13]$, $(13,23]$) to discover developmental factors. 
We compute the average intensity vectors among age groups and run the statistical-significance test on these results. Figure \ref{pic:stats-age} shows the 
statistical test results between age groups of all countries. 
We cannot reject the hypothesis about the originating distributions being equal for the age groups $(7,9]$ and $(9,11]$, as well as for the groups $(11,13]$ and $(13,23]$. 
This brings us to the conclusion that for the average intensity analysis it is enough to have three distinct groups $(1,7]$, $(7,11]$ and $(11,23]$, 
rather than multiple groups with hardly distinguishable properties. This result follows the theory of \cite{piaget:00}, who discovered that there is a significant shift 
in children's psychology around the age of seven. Around this age, the child's reasoning, which was preoperational, becomes operational. At around the age of twelve it 
becomes concrete operational and formal operational after twelve. This result was so influential that even the school education system was reformed to better 
correlate with the children's cognitive development \cite{wood:06}.

\subsubsection{Average intensity differences by task}
\label{sec:gravitytask}
As we mentioned in Section \ref{sec:data}, there are two distinct groups of drawings inside Switzerland: drawings for the task to represent God(s)  
and drawings that were obtained in the experiments with a more general task. We denoted these datasets as God(s) and General drawings. 
The two subsets were collected under similar conditions -- same country, age distribution and pencils. 
\begin{figure}[t]
  \centering
  \subfloat[Average intensity vector of the God(s) dataset.]{\includegraphics[width=0.4\columnwidth]{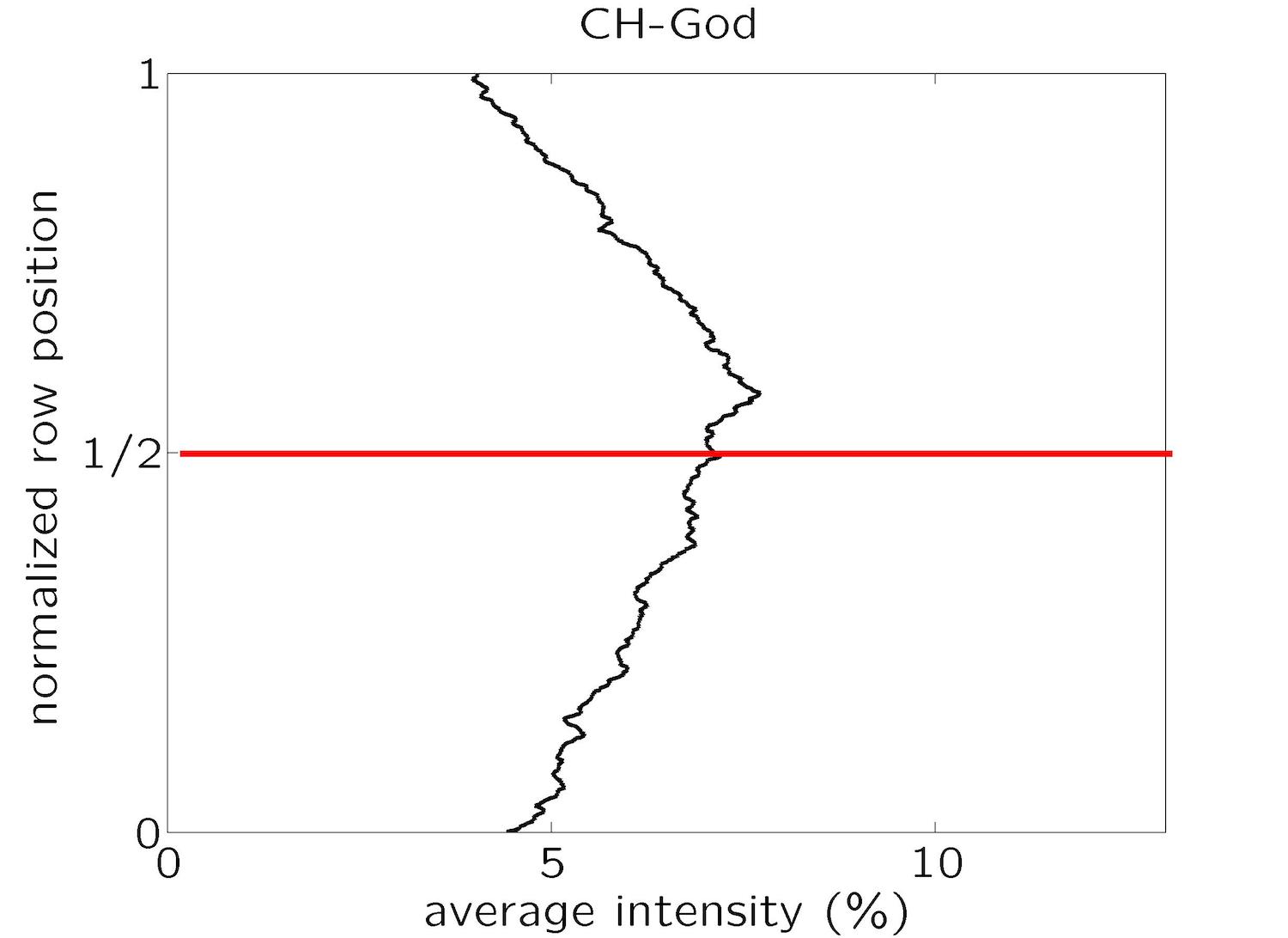}\label{pic:gravity-god}}\hspace{0.5in}
  \subfloat[Average intensity vector of the General dataset]{\includegraphics[width=0.4\columnwidth]{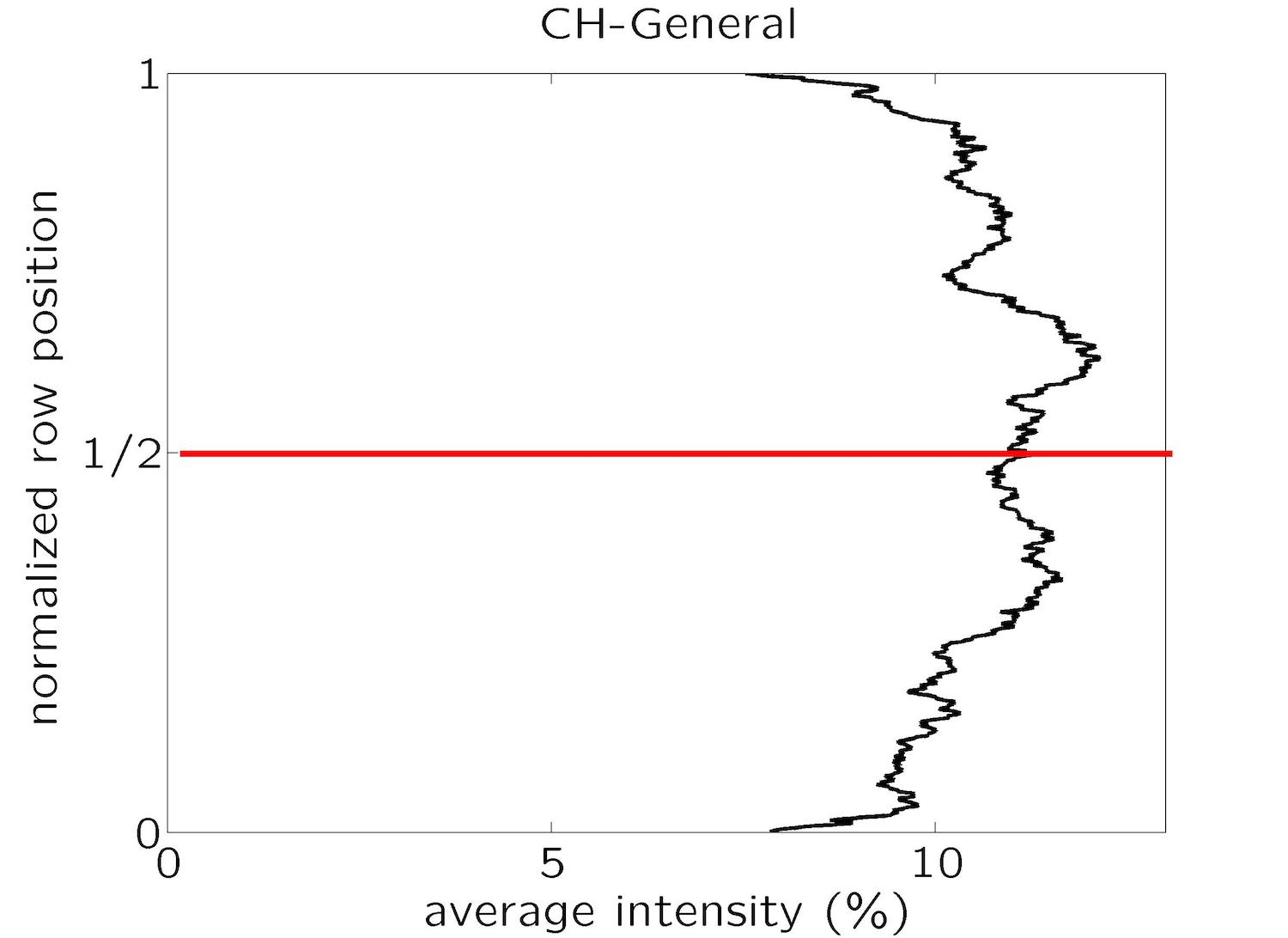}\label{pic:gravity-not-god}}
  \caption{Comparison of the average intensity vectors in subsets of drawings of God(s) and General drawings in Switzerland. The above distributions are statistically different according to our statistical test.} 
  \label{pic:gravity-god-not-god}
\end{figure}
Figure \ref{pic:gravity-god} shows the average intensity vector for the God(s) subset, and Figure \ref{pic:gravity-not-god} shows the resulting vector of the General subset.
We can clearly see the difference in the shape of the vectors and the gravity centers. If we compute the absolute proportion of the colored part of the drawings, 
we observe a difference in it as well: $31.6\%$ for the General subset versus $16.7\%$ for the God(s) drawings. Our samples of God(s) and General drawings from Switzerland are quite limited, 
this is why it is critical to check the statistical significance before making any conclusions. Applying the test, we easily reject the hypothesis of the distributions being 
equal at $95\%$ confidence level, even with our limited data. This is an important insight for researchers in the psychology of religion as it shows that the precise formulation 
of the task is extremely important in the data collection process.

\subsection{The Distribution of Green and Yellow}
\label{sec:intensitycolor}
Empirical evidence from looking at the drawings lets us believe that the colors green and yellow play an important role in the interpretation of children's drawings of God(s). 
We hypothesize that, in the task ``drawing of God(s)'', the color green is associated with nature and is used to characterize 
the more earthly location of the representation of God(s). On the other hand, we assume that the color yellow is associated with the representation of ``light''.  
In this section, we calculate how extensively each of these colors is used in the drawings and how their concentration is distributed on the drawings' area. 

To calculate color distributions on the scans, we first scale all images to size $k\times k$, in our case the choice of $k$ was $200$.
The resulting image is
\[
\mathbf{M}^{(i)} =
\begin{pmatrix}
  m_{1,1}^{(i)} & m_{1,2}^{(i)} & \cdots & m_{1,k}^{(i)} \\
  m_{2,1}^{(i)} & m_{2,2}^{(i)} & \cdots & m_{2,k}^{(i)} \\
  \vdots  & \vdots  & \ddots & \vdots  \\
  m_{k,1}^{(i)} & m_{k,2}^{(i)} & \cdots & m_{k,k}^{(i)}
\end{pmatrix}, \quad 
i=1,...,N,
\] 
where each $m_{j,l}^{(i)} \in \mathbb{R}^{1 \times 3}, \; j = 1,\dots,k, \; l =
1,\dots,k$ represents a color in the CIE LAB color space. We introduce a vector with representative values for the colors green and yellow and a distance threshold. 
The threshold defines how far a pixel should be from the model vector in order to be perceived as green or yellow. 
We define the green vector as $[50, -50, 50]$ and the yellow vector as $[80, 0, 100]$. These values were selected empirically to correspond to an intuitive human perception 
of the colors green and yellow, respectively. The distance threshold is fixed to $50$, which enables us to distinguish different color hues from different colors altogether. 
We define a distance matrix $\mathbf{D} \in \mathbb{R}^{k \times k}$ from each pixel in the image to the color of interest. 
\begin{figure}[h]
  \centering
  \subfloat[Intensity of the color green.]{\includegraphics[width=0.25\columnwidth]{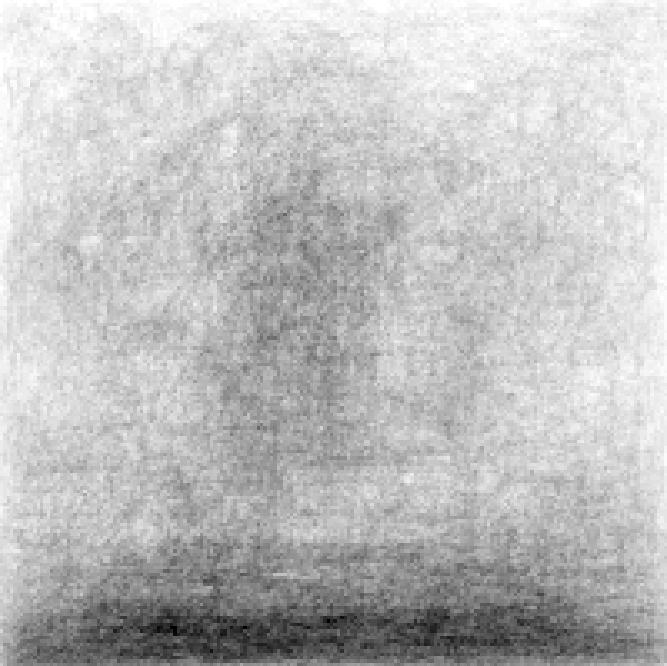}\label{pic:yellow}}\hspace{0.5in}
  \subfloat[Intensity of the color yellow.]{\includegraphics[width=0.25\columnwidth]{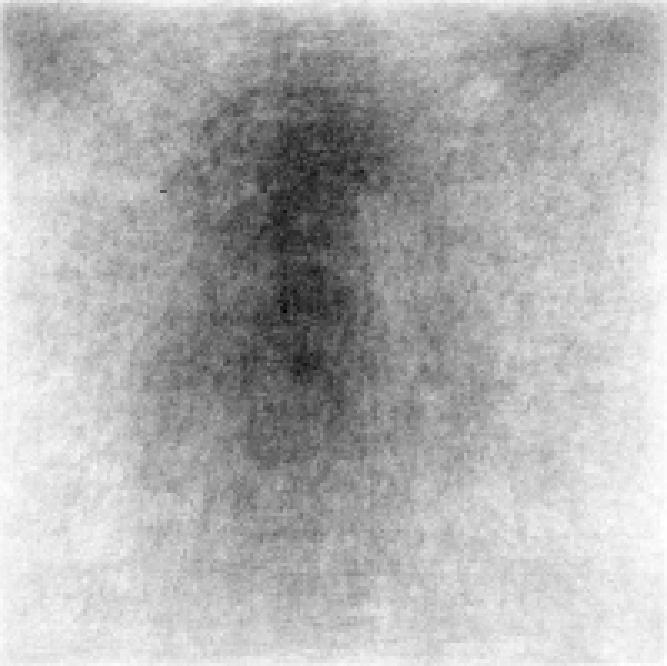}\label{pic:green}}
  \caption{Intensity of the colors green and yellow in the dataset of drawings of God(s).
  The darker the point is, the more green and yellow pixels are present at this location in the drawings of the dataset.} 
  \label{pic:greenyellow}
\end{figure}
By applying the above threshold, we get a binary image, where $1$ corresponds to distances less than or equal to the defined threshold, and $0$ to distances larger than the threshold. 
In other words, $1$ is an approximation of the presence of green (or yellow) and $0$ is its absence. Figure \ref{pic:greenyellow} shows the distribution of colors 
on the scans for green and yellow pixels in the drawings. The darker the point in the figure, the more pixels of a particular color (green or yellow, respectively) 
are present. We investigate the concentration of the colors green and yellow in the drawings for different regions and age groups. 


\subsubsection{Color analysis by country}
\label{sec:coloranalysiscountry}
\begin{figure}[t]
  \centering
  \subfloat[CH]{\includegraphics[width=0.18\columnwidth]{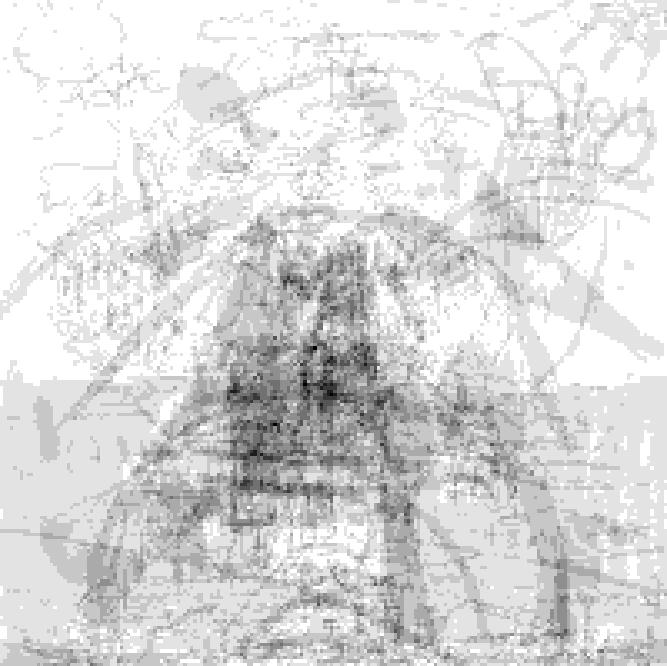}}\hspace{0.1in}
  \subfloat[JP]{\includegraphics[width=0.18\columnwidth]{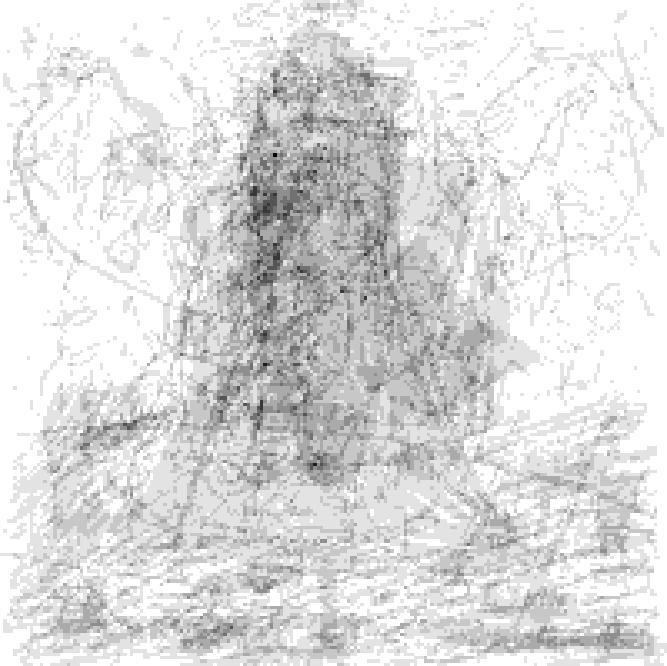}}\hspace{0.1in}
  \subfloat[RO]{\includegraphics[width=0.18\columnwidth]{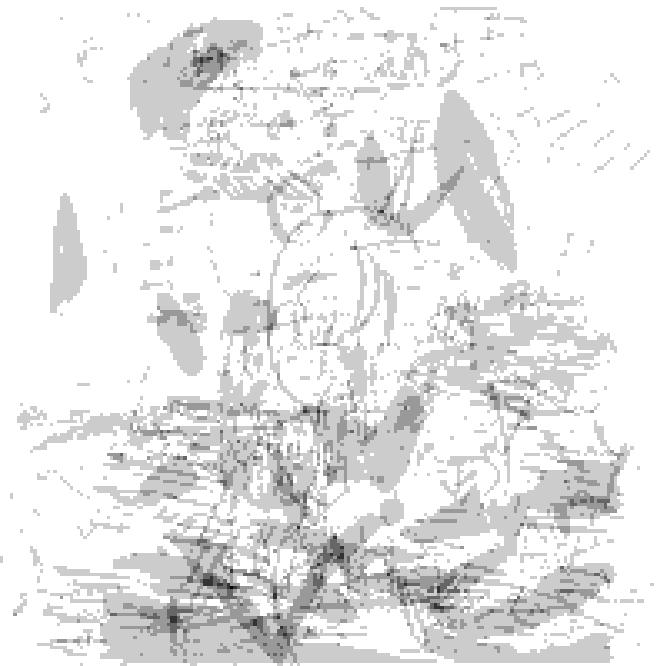}}\\
  \subfloat[RU-bo]{\includegraphics[width=0.18\columnwidth]{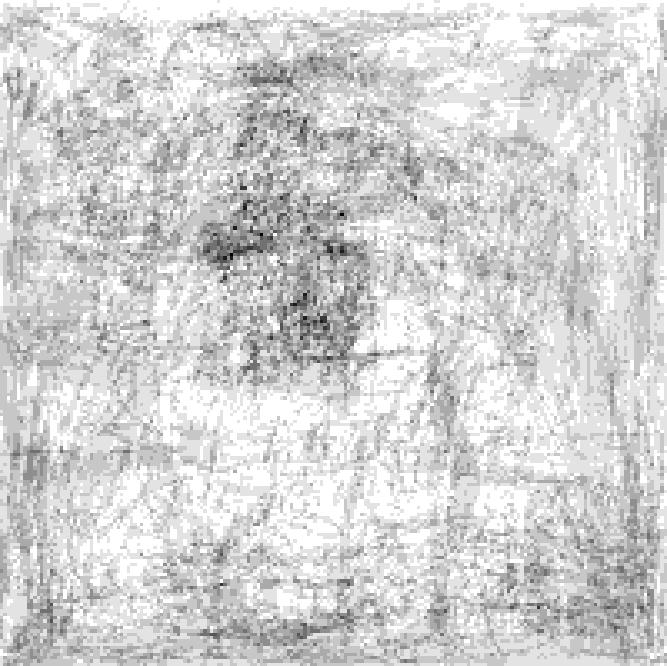}}\hspace{0.1in}
  \subfloat[RU-sp]{\includegraphics[width=0.18\columnwidth]{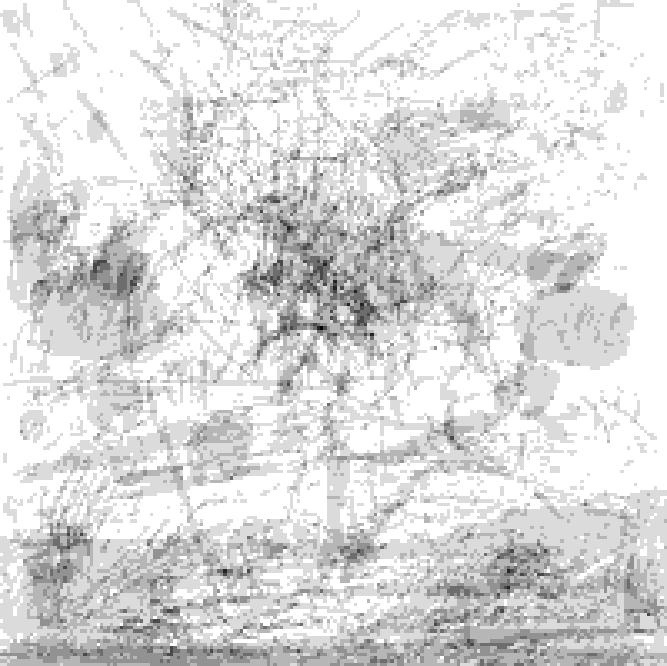}}\hspace{0.1in}
  \subfloat[US]{\includegraphics[width=0.18\columnwidth]{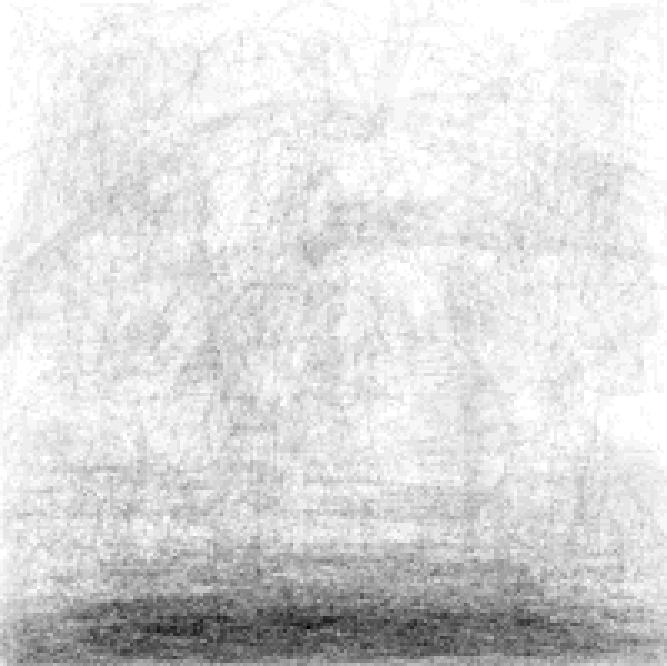}}
  \caption{Intensity of the color \emph{green} among different countries in the dataset. The darker the point is, the more green pixels are present at this location in the drawings of the dataset.} 
  \label{pic:greencountry}
\end{figure}

Figure \ref{pic:greencountry} illustrates the distribution, grouped by countries, of the color green in the drawings. Observing the resulting intensities, 
we can make a few conclusions. For example, in Japan the color green is more often used for the central object. If we compare the two Russian regions, we notice that 
in the drawings from Buryatia green pixels are more equally spread over the drawing than in those from Saint-Petersburg. We can explain this phenomenon with the fact that in 
Buryatia there exists a more shamanic conception where God(s) is associated with spirits living everywhere around humans. 
We see that these children put their supernatural agents in the context of the earth. 
US children definitely associate green with grass on the ground and it shows a strong stereotype about how they learn to draw.
Even though we cannot easily make conclusions with this limited dataset and the patterns do not look easy to analyze, we can still see 
some evidence about the cultural differences in the countries, which gives us ideas for new hypotheses.
\begin{figure}[t]
  \centering
  \subfloat[CH]{\includegraphics[width=0.18\columnwidth]{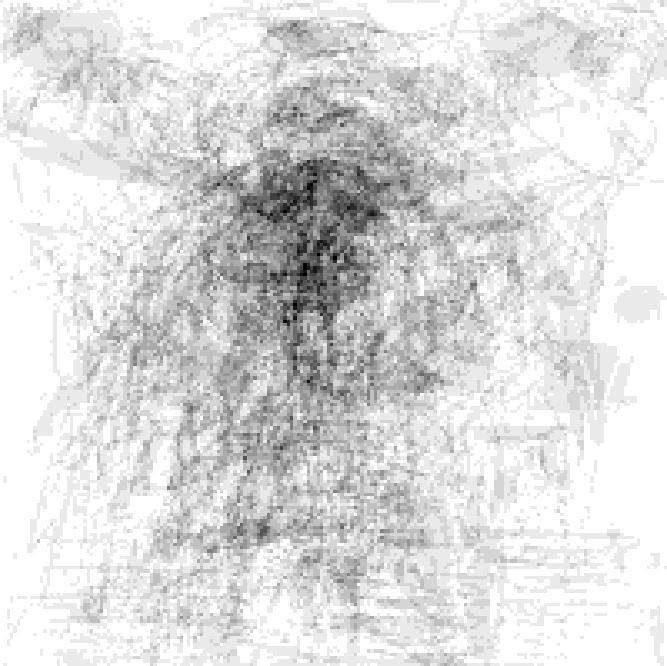}}\hspace{0.1in}
  \subfloat[JP]{\includegraphics[width=0.18\columnwidth]{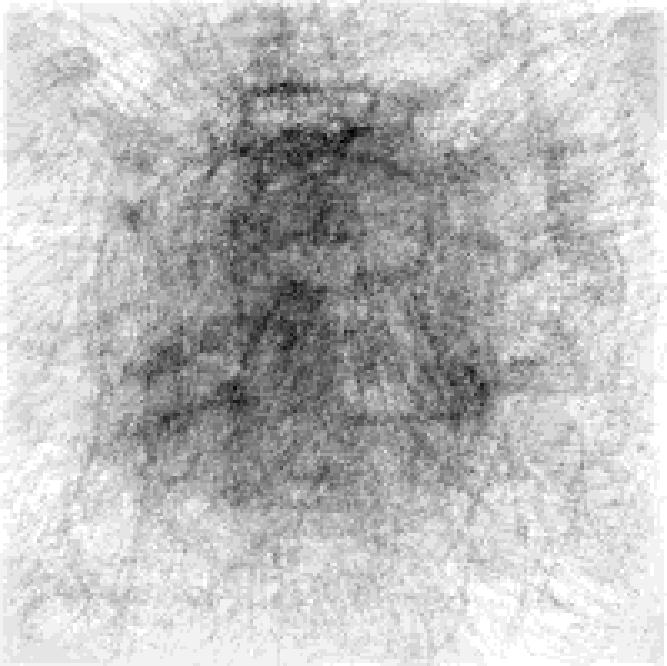}}\hspace{0.1in}
  \subfloat[RO]{\includegraphics[width=0.18\columnwidth]{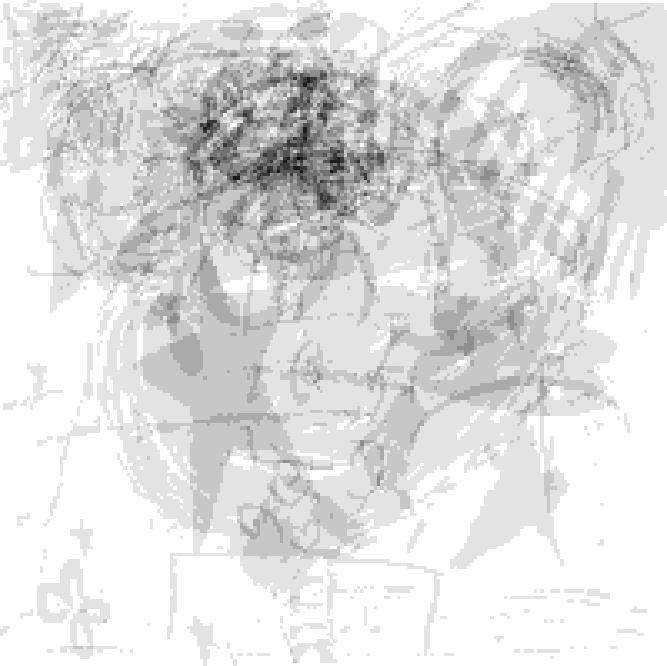}}\\
  \subfloat[RU-bo]{\includegraphics[width=0.18\columnwidth]{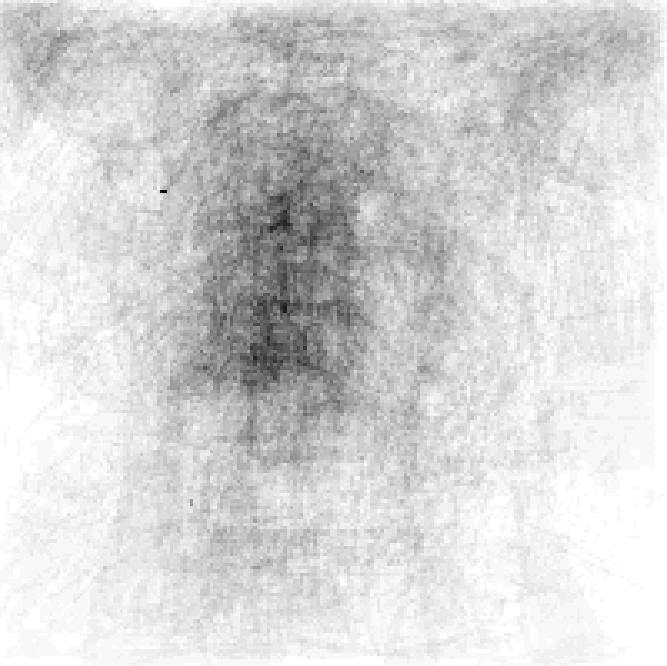}}\hspace{0.1in}
  \subfloat[RU-sp]{\includegraphics[width=0.18\columnwidth]{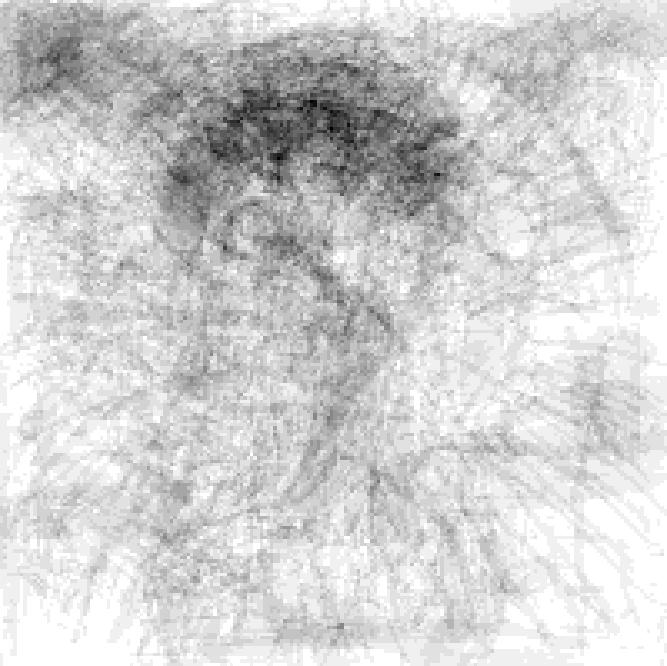}}\hspace{0.1in}
  \subfloat[US]{\includegraphics[width=0.18\columnwidth]{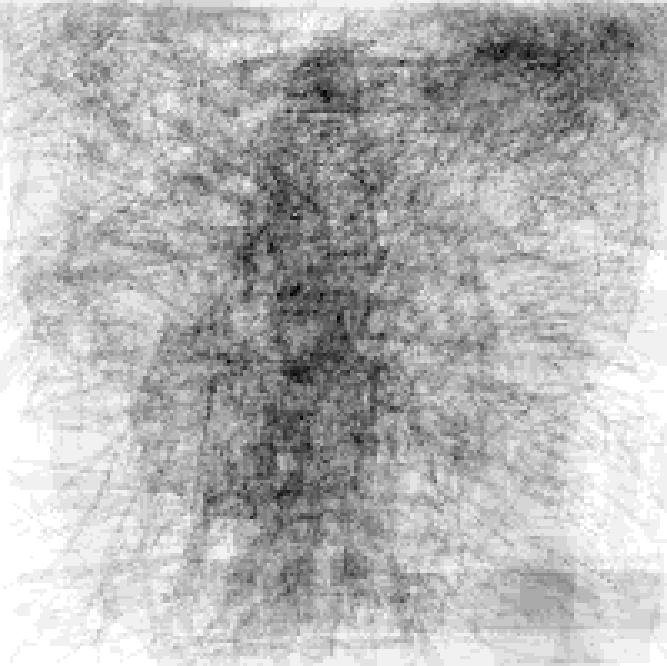}}
  \caption{Intensity of the color \emph{yellow} among different countries in the dataset. The darker the point is, the more yellow pixels are present at this location in the drawings of the dataset.} 
\label{pic:yellowcountry}
\end{figure}

\begin{figure}[t]
  \centering
  \subfloat[Intensity of the color yellow in Saint-Petersburg.]{\includegraphics[height=0.27\columnwidth]{yellow-ru-sp.jpg}}\hspace{0.2in}
  \subfloat[Drawing from Saint-Petersburg.]{\fbox{\includegraphics[height=0.27\columnwidth]{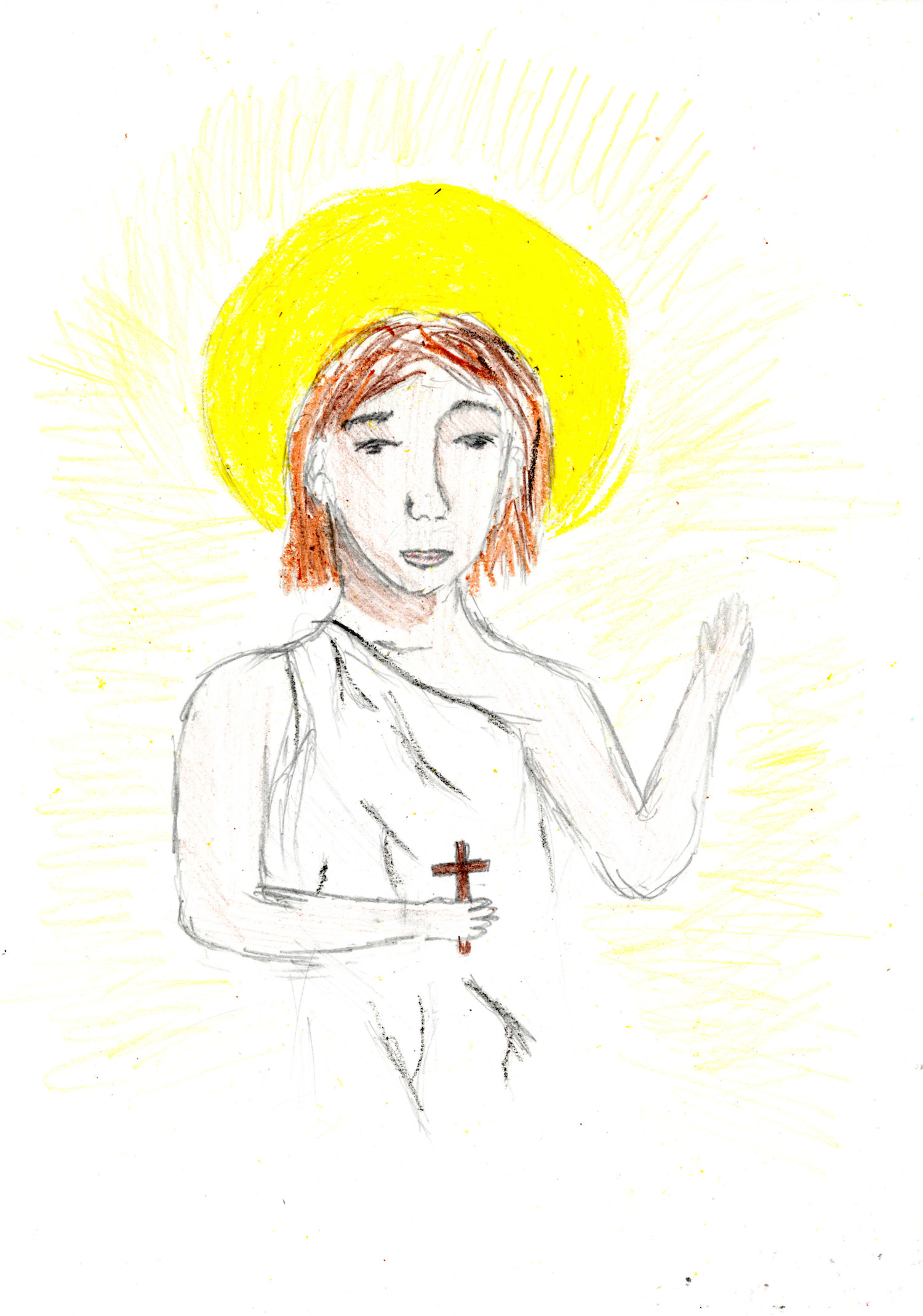}}}
  \caption{Illustration of the influence of the Orthodox church on children's drawings in Saint-Petersburg.} 
  \label{pic:stpetersburg}
\end{figure}

Figure \ref{pic:yellowcountry} shows the distribution of the color yellow in different countries. 
It is obvious from the figure that the color yellow is often associated with a central object. 
Sometimes yellow is a sign of presence of a ``sun" object in a drawing in the upper left or right corner of the page.
We discover an interesting pattern in the Saint-Petersburg region (also in Romania, but to a smaller degree) -- the location of the yellow cloud is 
shifted to the upper part of the image and has a horizontal oval shape. By directly inspecting the drawings, we find that there is a common strategy   
in the Orthodox tradition causing this effect. The pattern is illustrated in Figure \ref{pic:stpetersburg}: the drawing of a person with an aura clearly 
illustrates the influence of the Orthodox church on children's representation of God(s) in Saint-Petersburg. Children adapted the church's art traditions from popular icons 
that are very frequent in Orthodox churches. We also observe some other cultural particularities in the distribution of the color yellow.
\begin{figure}[t]
  \centering
  \includegraphics[height=0.4\columnwidth]{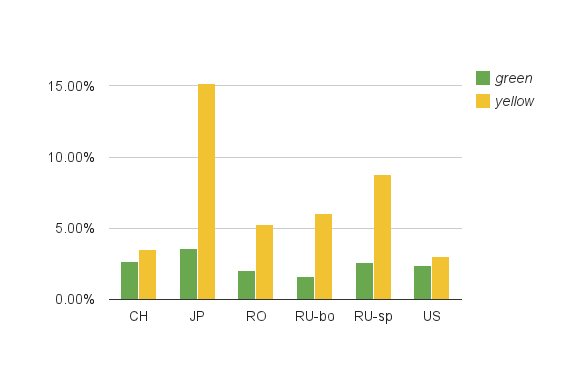}
  \caption{Proportion of the colors green and yellow of different countries in the dataset.} 
  \label{pic:greenyellowcountry}
\end{figure}
The pattern in Switzerland is more similar to other Christian cultures, such as those in Russia or Romania, rather than those in Japan. 
In Japan, we observe a huge yellow cloud in the center of the drawings: when looking at the drawings, we observe that it corresponds to a yellow figure or rays of light around it. 

Figures \ref{pic:greencountry} and \ref{pic:yellowcountry} do not give us any information about the absolute intensities of the retrieved colors, 
they only demonstrate the comparative spread of the colors and do not allow us to make a quantitative comparison between these two important colors. 
To obtain some information about absolute values, we compute the portion of the colors green and yellow in the drawings. 
The results are shown in Figure \ref{pic:greenyellowcountry}. We notice that, for example, Japanese children use the color yellow more frequently, 
and in all countries the color yellow dominates over green, even though marginally in Switzerland and in the USA.  
\begin{figure}[t]
  \centering
  \subfloat[$(1,7{]}$]{\includegraphics[width=0.18\columnwidth]{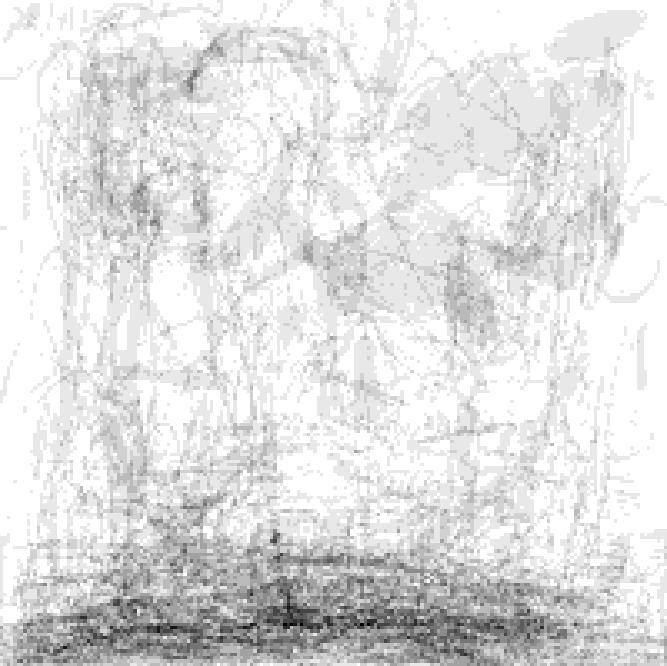}}\hspace{0.1in}
  \subfloat[$(7,9{]}$]{\includegraphics[width=0.18\columnwidth]{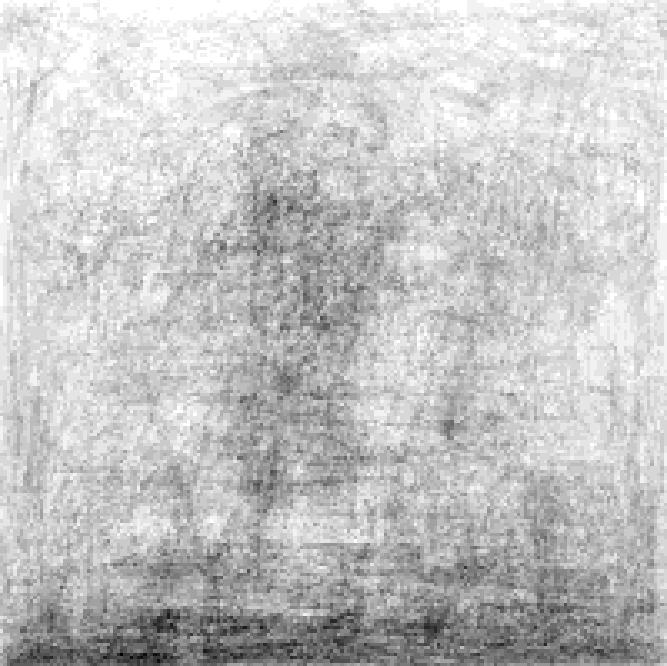}}\hspace{0.1in}
  \subfloat[$(9,11{]}$]{\includegraphics[width=0.18\columnwidth]{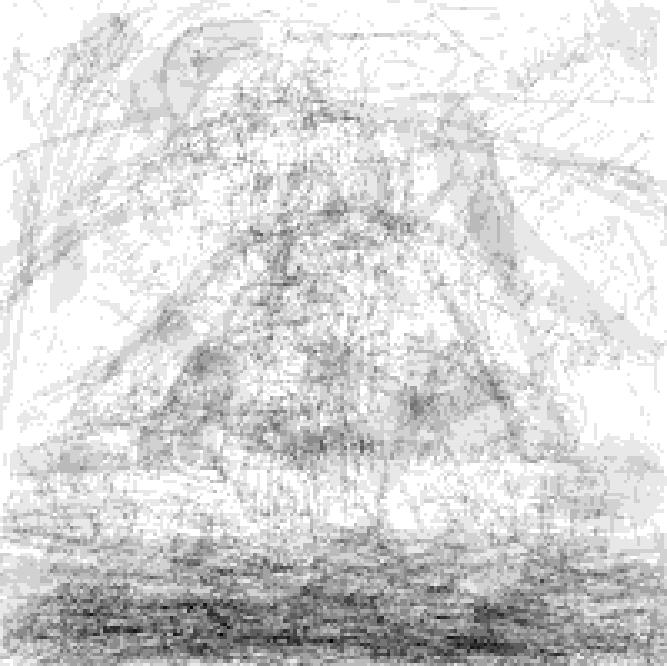}}\hspace{0.1in}
  \subfloat[$(11,13{]}$]{\includegraphics[width=0.18\columnwidth]{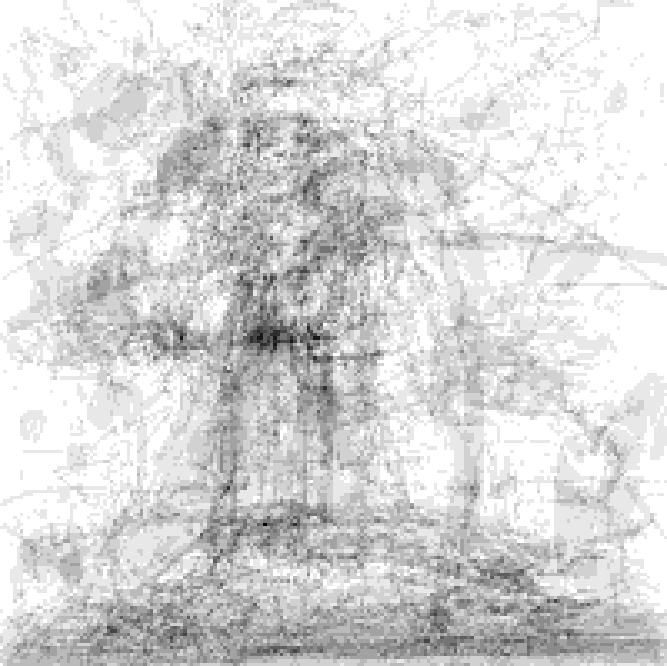}}\hspace{0.1in}
  \subfloat[$(13,23{]}$]{\includegraphics[width=0.18\columnwidth]{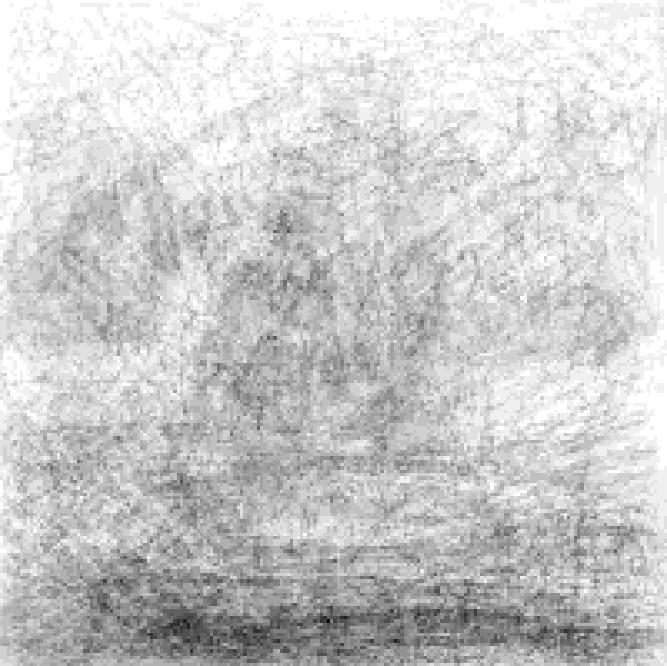}}\hspace{0.1in}
  \caption{Intensity of the color \emph{green} among different age groups in the dataset. The darker the point is, the more green pixels are present at this location in the drawings of the dataset.} 
  \label{pic:greenage}
\end{figure}
\begin{figure}[t]
  \centering
  \subfloat[$(1,7{]}$]{\includegraphics[width=0.18\columnwidth]{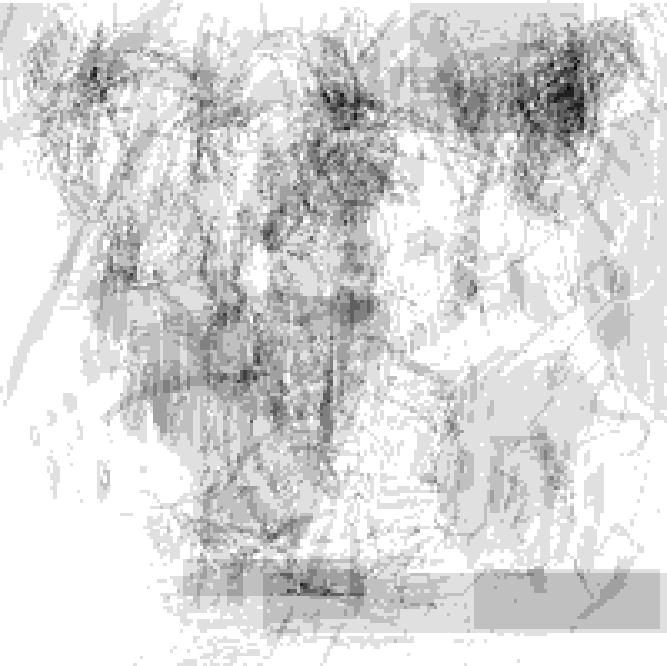}}\hspace{0.1in}
  \subfloat[$(7,9{]}$]{\includegraphics[width=0.18\columnwidth]{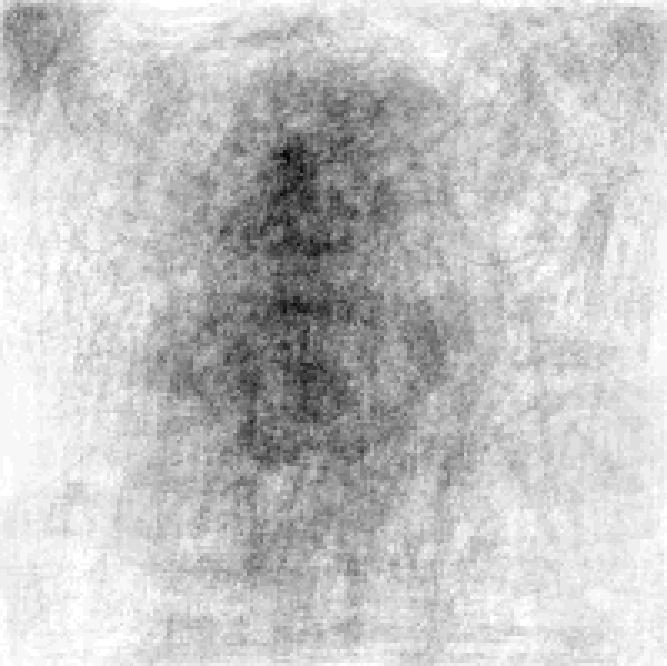}}\hspace{0.1in}
  \subfloat[$(9,11{]}$]{\includegraphics[width=0.18\columnwidth]{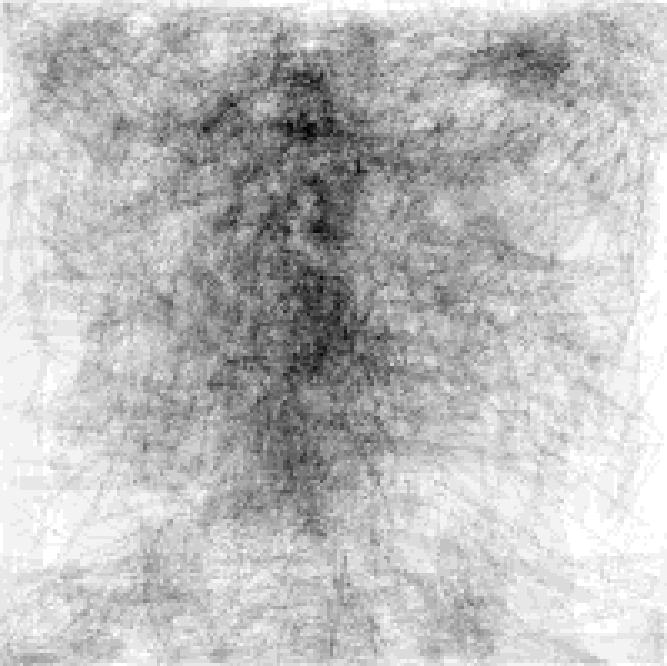}}\hspace{0.1in}
  \subfloat[$(11,13{]}$]{\includegraphics[width=0.18\columnwidth]{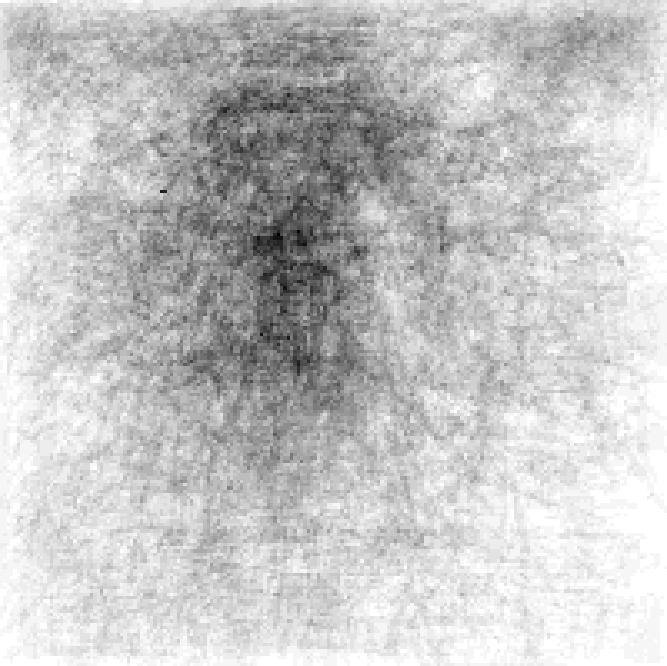}}\hspace{0.1in}
  \subfloat[$(13,23{]}$]{\includegraphics[width=0.18\columnwidth]{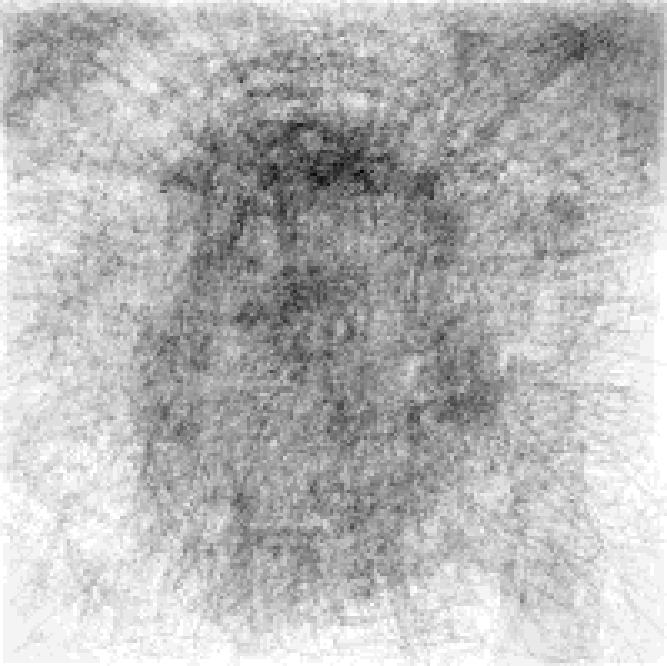}}\hspace{0.1in}
  \caption{Intensity of the color \emph{yellow} among different age groups in the dataset. The darker the point is, the more yellow pixels are present at this location in the drawings of the dataset.} 
  \label{pic:yellowage}
\end{figure}

\subsubsection{Color analysis by age}
\label{sec:coloranalysisage}
\begin{figure}[t]
  \centering
  \includegraphics[height=0.4\columnwidth]{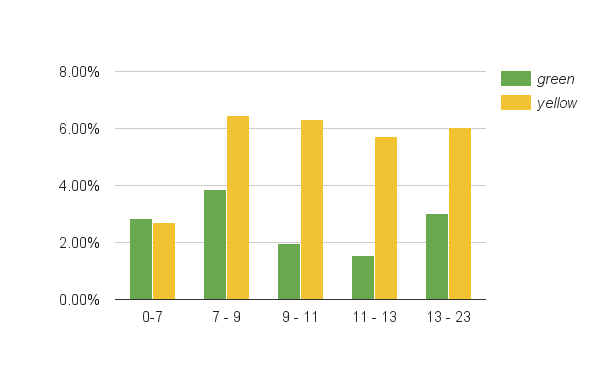}
  \caption{Proportion of the colors green and yellow of different age groups in the dataset.} 
  \label{pic:greenyellowage}
\end{figure}

We also investigate the distribution of colors for different age groups. Figures \ref{pic:greenage} and \ref{pic:yellowage} show the distribution 
of the colors green and yellow, respectively, for the age groups $(1,7]$, $(7,9]$, $(9,11]$, $(11,13]$ and $(13,23]$. In Figure \ref{pic:greenyellowage}, we report the absolute 
intensities of these colors compared to the number of colored pixels. Combining the evidence from these two types of figures, we notice that young children, in the 
age group $(1,7]$, behave differently from other children in their usage of yellow. First, we see that the proportion of the color yellow in young children's drawings is half as 
much as the proportion of yellow in any other age group. For young children, the color yellow is used for drawing a sun, which represents light, but only as part of a landscape. 
It is very common among children to draw a sun as an element of their drawing \cite{hargreaves:81}. 
In this case, it is not related to the concept of God(s) and occupies only a small part of the entire drawing area: a little element in a corner or in the top of the drawing. 

Comparing this age group with the rest, we observe a significant difference in the drawing strategy: older children associate yellow more with the main subject -- God(s). 
One of the reasons could be that the association of God(s) with light is learned by children with age and religious education. 
The distribution of the two colors (green and yellow) among other age groups seems to slightly vary, however, drawings of children in the group $(1,7]$ are 
noticeably distinct (see Figures \ref{pic:greenage}, \ref{pic:yellowage}). 
The proportion of the colors green and yellow supports the hypothesis of variations in the presence of earthly/nature and light features in the representation of God(s). 
We hypothesize that the tendency to put more yellow into the drawings after the age of seven can be interpreted as an attempt to create an image of God(s) who is closer to the light. 
Following our hypothesis, the color yellow seems to be a \emph{general code} among children for representing light in the task ``drawing of God(s)''. On the other hand, the color green 
is a less general and systematic drawing code than the color yellow. For example, Japanese children use the color green more for the main figure than for elements of nature.  

\subsubsection{Color analysis by task}
\label{sec:intensitygodvsnongod}
\begin{figure}[t]
  \centering
  \subfloat[Green in God(s).]{\includegraphics[height=0.18\columnwidth]{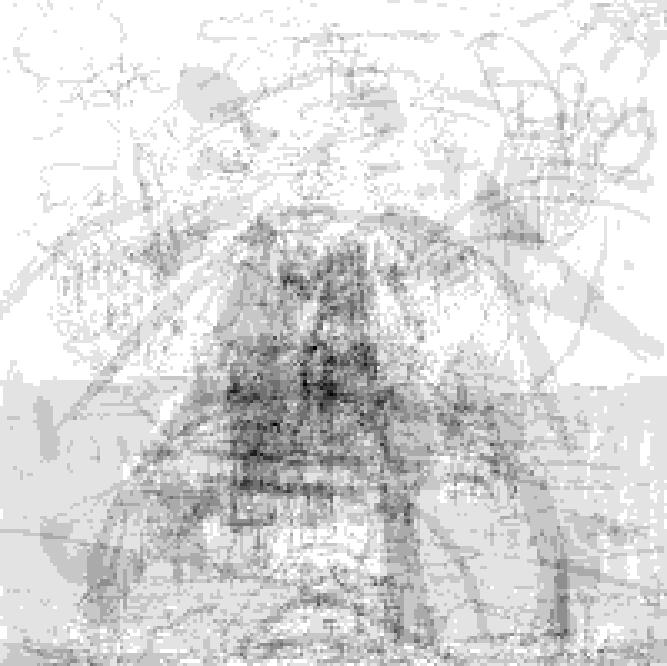}}\hspace{0.1in}
  \subfloat[Yellow in God(s).]{\includegraphics[height=0.18\columnwidth]{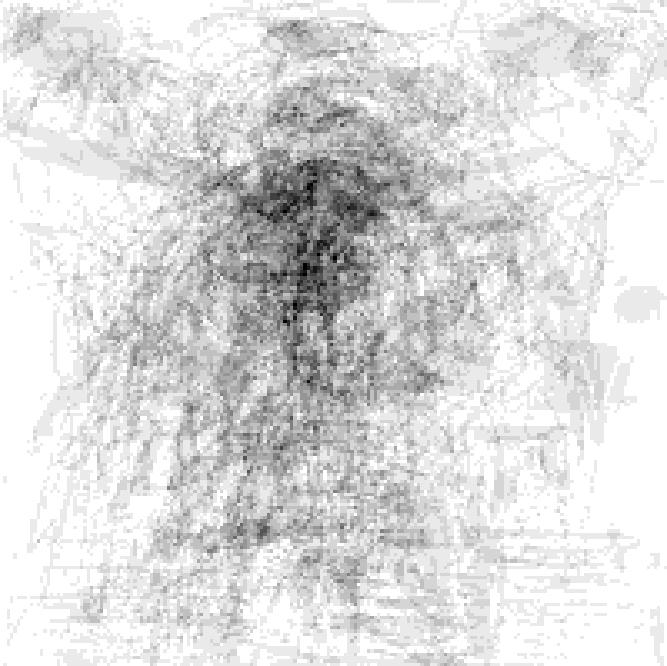}}\hspace{0.1in}\\
  \subfloat[Green in General.]{\includegraphics[height=0.18\columnwidth]{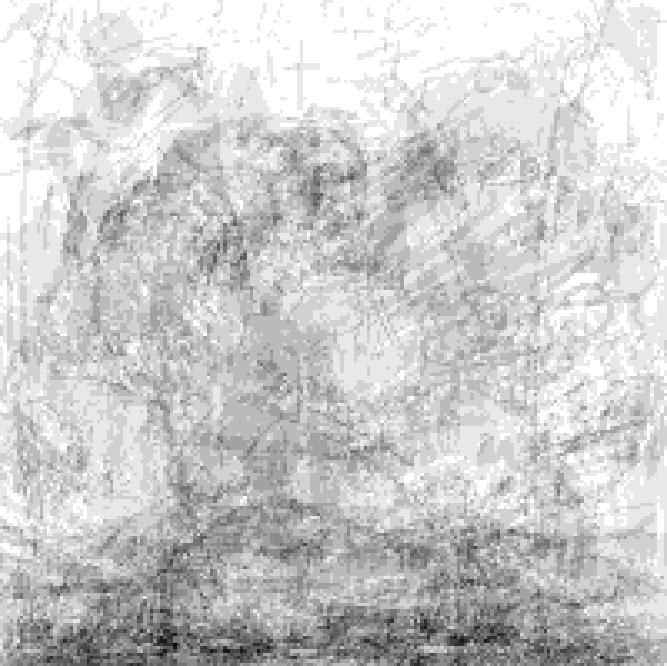}}\hspace{0.1in}
  \subfloat[Yellow in General.]{\includegraphics[height=0.18\columnwidth]{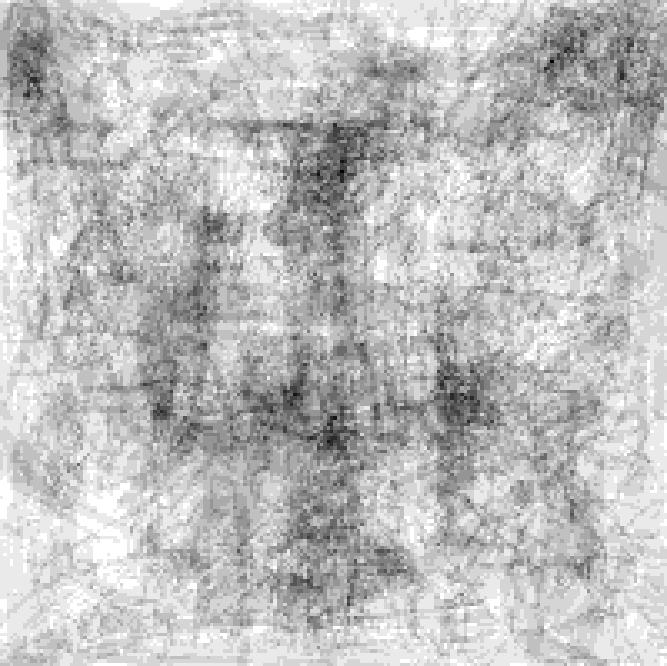}}\hspace{0.1in}
  \caption{Intensities of the colors green and yellow in the God(s) and General drawings. 
  The difference in the task formulation is reflected in the way children use these two colors.} 
  \label{pic:greenyellowgodgeneral}
\end{figure}

To check how the strategy of drawing depends on the task the children were given, Figure \ref{pic:greenyellowgodgeneral} compares the distribution of the colors green and yellow  
for the God(s) and General subsets in Switzerland. 
We observe that, in the God(s) subset, green and yellow are concentrated in the center, thus corresponding to a main object. 
In the General drawings, however, the color green is concentrated in the bottom of the drawing and the color yellow in the center and the corners. 
This difference indicates that the children's approach to address the task significantly depends on the formulation.

\subsection{Palette Extraction}
\label{sec:paletteanalysis}
In order to analyze the variety of colors that are used in drawings of God(s), we have to identify the pencils that the children used. 
For this task, we develop a procedure for palette extraction based on segmentation. We then present and discuss the results for different categories.

\subsubsection{Palette extraction method}
There are several image segmentation methods in computer vision, such as Mean shift \cite{comaniciu:02}, Normalized cuts \cite{shi:00}, 
K-means and Gaussian mixture models. We decided to use the K-means algorithm \cite{macqueen:67} for this task, because it is simple to implement, it is fast, 
it always converges and operates on Euclidean distances that reflect our similarity perception in the CIE LAB space.
\begin{figure}[h]
  \begin{minipage}[t]{0.5\linewidth}
    \centering
    \includegraphics[height=0.6\columnwidth]{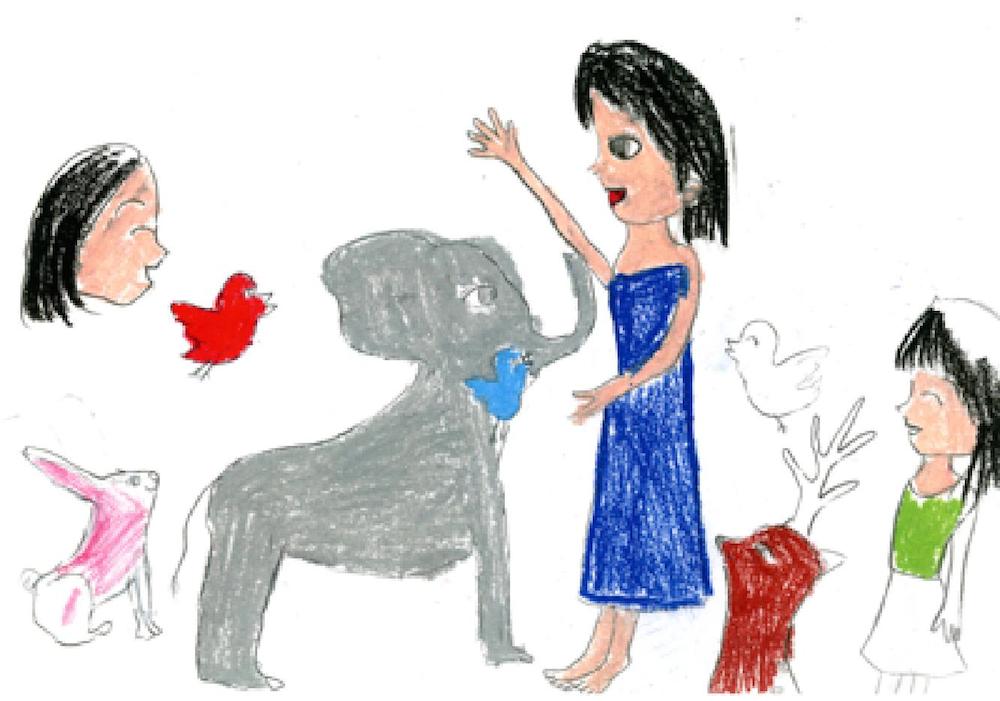}
  \end{minipage}
  \begin{minipage}[t]{0.5\linewidth}
    \centering
    \includegraphics[height=0.6\columnwidth]{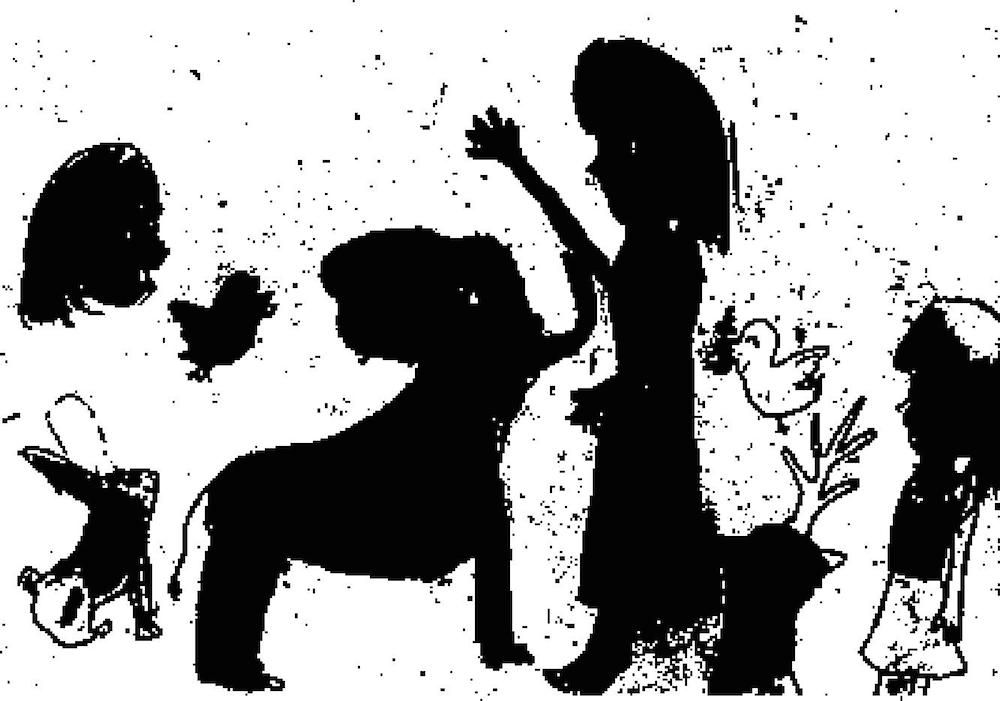}
  \end{minipage}
  \caption{A drawing from the dataset and its extracted foreground.} 
  \label{pic:background}
\end{figure} 

In order to take into account only colored pixels in our segmentation method, we need to separate the colored part of the drawing 
(\emph{colored foreground}) from the part that is not drawn (\emph{background}). 
The goal of colored foreground extraction is to separate the white paper background of the scanned images from the colored foreground. This separation increases 
the robustness of the segmentation, because a substantial amount of noise is removed from the data after the colored foreground extraction process.
For this task, we use a \emph{Gaussian mixture model (GMM)} with two mixture components. To initialize the mixture components, we take advantage of our prior knowledge: 
we know that plain paper has an $L$ channel value higher than colored pixels in the CIE LAB color space. 
Moreover, background usually occupies most of the area of the paper. This prior knowledge implies a reliable initialization of our mixture components, which leads 
to rapid convergence of the \emph{Expectation Maximization} algorithm \cite{dempster:77} that is used to train the model. 
Figure \ref{pic:background} shows an image from our dataset, together with its extracted colored foreground in black. The resulting components have a mixing proportion of 
$0.62$ and $0.38$ and their average $L$ channel values are $254.9977$ and $166.3169$ for background and colored foreground, respectively. The result coincides with our initial 
assumption about the amount of colored foreground pixels and the difference in the average $L$ value. 
The chosen method is appropriate for any type of image, because there is no need to adapt parameters depending on different brightness levels and white balance of scans. 
\begin{figure}[t]
  \centering
  \includegraphics[height=0.4\columnwidth]{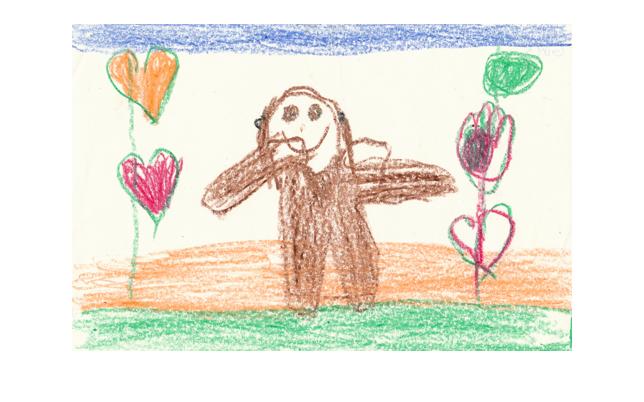}
  \caption{Original drawing from the dataset used for the palette extraction experiment.} 
  \label{pic:originalforpalette}
\end{figure}
\begin{figure}[t]
  \begin{minipage}[t]{0.24\linewidth}
    \centering
    \includegraphics[height=0.25\columnwidth]{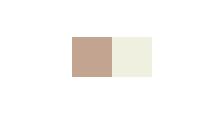}
  \end{minipage}
  \begin{minipage}[t]{0.24\linewidth}
    \centering
    \includegraphics[height=0.25\columnwidth]{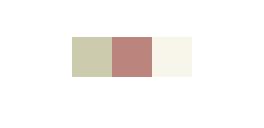}
  \end{minipage}
  \begin{minipage}[t]{0.24\linewidth}
    \centering
    \includegraphics[height=0.25\columnwidth]{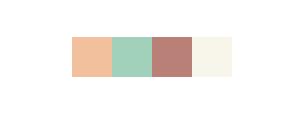}
  \end{minipage}
  \begin{minipage}[t]{0.24\linewidth}
    \centering
    \includegraphics[height=0.25\columnwidth]{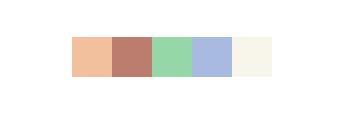}
  \end{minipage} \\
  \subfloat[$2$ clusters.]{\includegraphics[width=0.24\columnwidth]{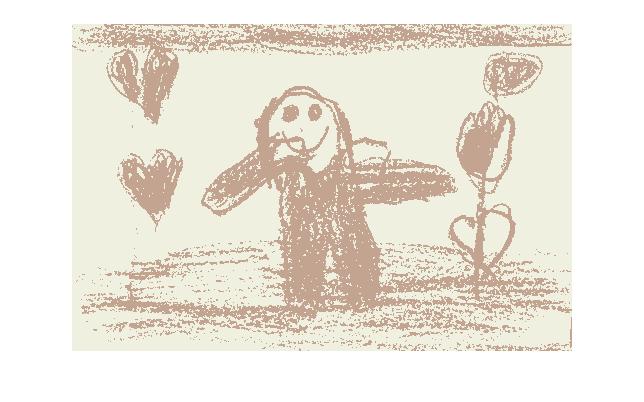}}
  \subfloat[$3$ clusters.]{\includegraphics[width=0.24\columnwidth]{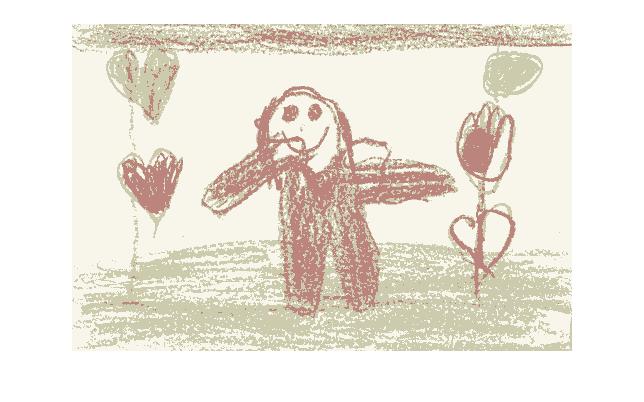}}
  \subfloat[$4$ clusters.]{\includegraphics[width=0.24\columnwidth]{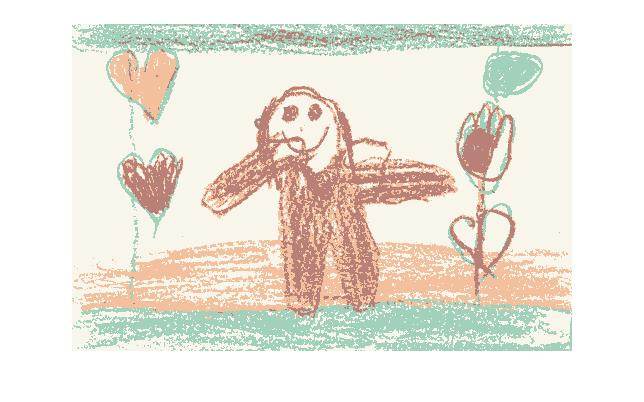}}
  \subfloat[$5$ clusters.]{\includegraphics[width=0.24\columnwidth]{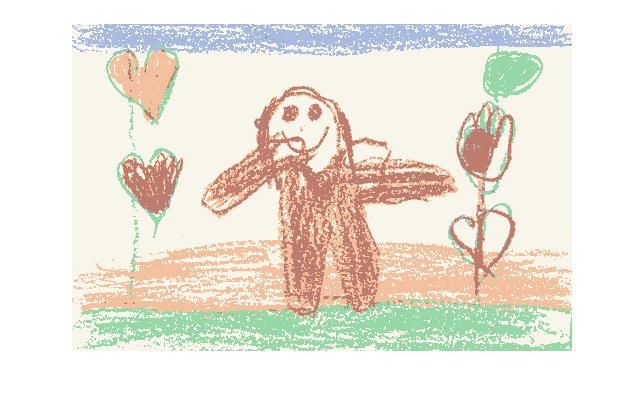}} \\
  \begin{minipage}[t]{0.24\linewidth}
    \centering
    \includegraphics[height=0.25\columnwidth]{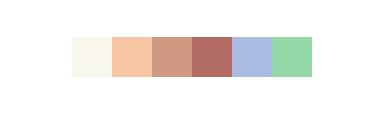}
  \end{minipage}
  \begin{minipage}[t]{0.24\linewidth}
    \centering
    \includegraphics[height=0.25\columnwidth]{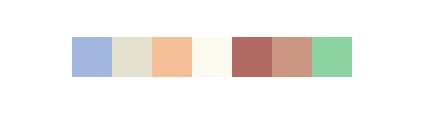}
  \end{minipage}
  \begin{minipage}[t]{0.24\linewidth}
    \centering
    \includegraphics[height=0.25\columnwidth]{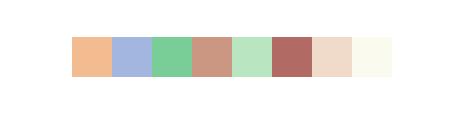}
  \end{minipage}
  \begin{minipage}[t]{0.24\linewidth}
    \centering
    \includegraphics[height=0.25\columnwidth]{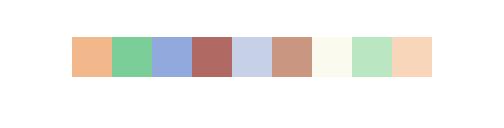}
  \end{minipage} \\
  \subfloat[$6$ clusters.]{\includegraphics[width=0.24\columnwidth]{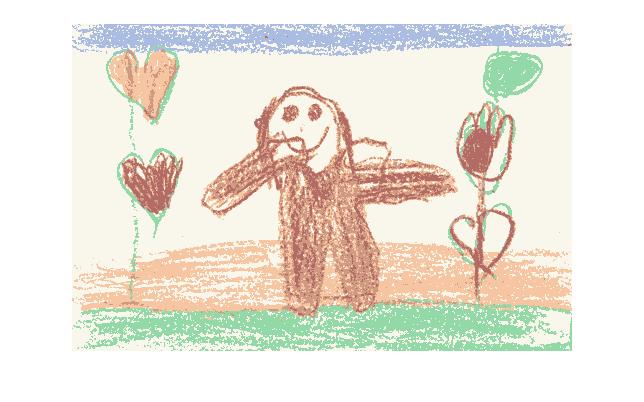}}
  \subfloat[$7$ clusters.]{\includegraphics[width=0.24\columnwidth]{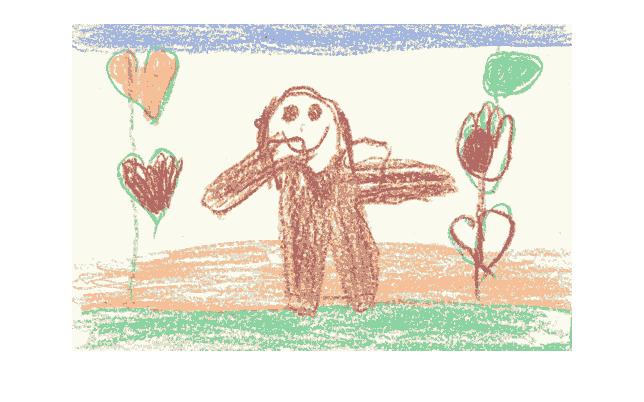}}
  \subfloat[$8$ clusters.]{\includegraphics[width=0.24\columnwidth]{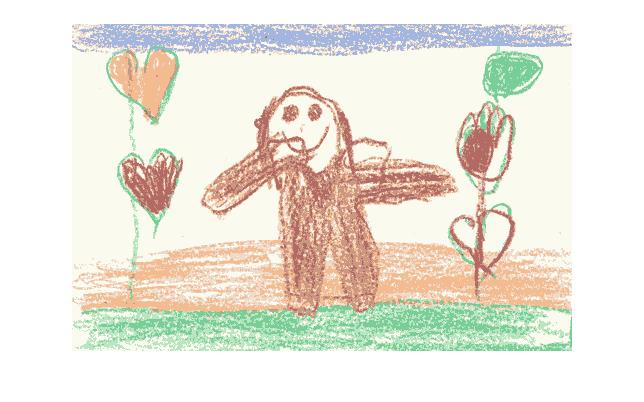}}
  \subfloat[$9$ clusters.]{\includegraphics[width=0.24\columnwidth]{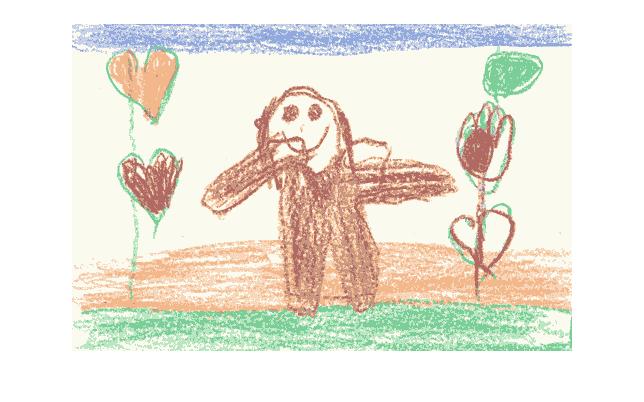}}
  \caption{Color palettes extracted from the drawing of Figure \ref{pic:originalforpalette} and its approximations. In the drawing approximations, we replace each pixel of the original scan by the centroid of the cluster it belongs to.} 
  \label{pic:approximation}
\end{figure}

Our input to the segmentation problem for each scan is a matrix of colored foreground pixels denoted by $\mathbf{M}_1 \in \mathbb{R}^{f \times 3}$. 
We cluster the $3$-D data into a specified number of groups, which varies from $2$ to $10$. 
Having determined the cluster centroids, we construct a drawing approximation, where all pixels are replaced by their cluster means and we investigate the result of the reconstruction. 
In this section, to demonstrate the complete palette reconstruction procedure, we use as an example the drawing in Figure \ref{pic:originalforpalette}.  
For the purpose of illustration, we also show the background. Figure \ref{pic:approximation} shows the extracted palettes and drawing approximations obtained with 
different numbers of clusters. 
To construct a drawing approximation, we replace each pixel in the scan by the centroid of the cluster it belongs to.
 
An obvious problem that arises from this method is the need to determine a priori the number of clusters. 
As we can see from Figure \ref{pic:approximation}, the quality of the approximation increases as we increase the number of clusters. 
Therefore, we cannot use the approximation quality as a measure of the quality of the clustering. 
However, we notice that when we move from four to five clusters, we begin to distinguish between the green grass and the blue sky, which is a critical distinction. 
When we introduce another extra cluster and use six centroids, we simply add another variety of brown. 
This addition does not bring much benefit to the reconstruction of pencil colors, because these colors usually correspond to lighter or harder pressure of the pencil. 
Therefore, we believe that a drawing can be accurately approximated by a limited number of pencils, without significant differences perceived by the 
human observer and we separate the drawing into clusters with the same limited number of colors.
\begin{figure}[t]
  \centering
  \subfloat[Silhouette measure.]{\includegraphics[width=0.4\textwidth]{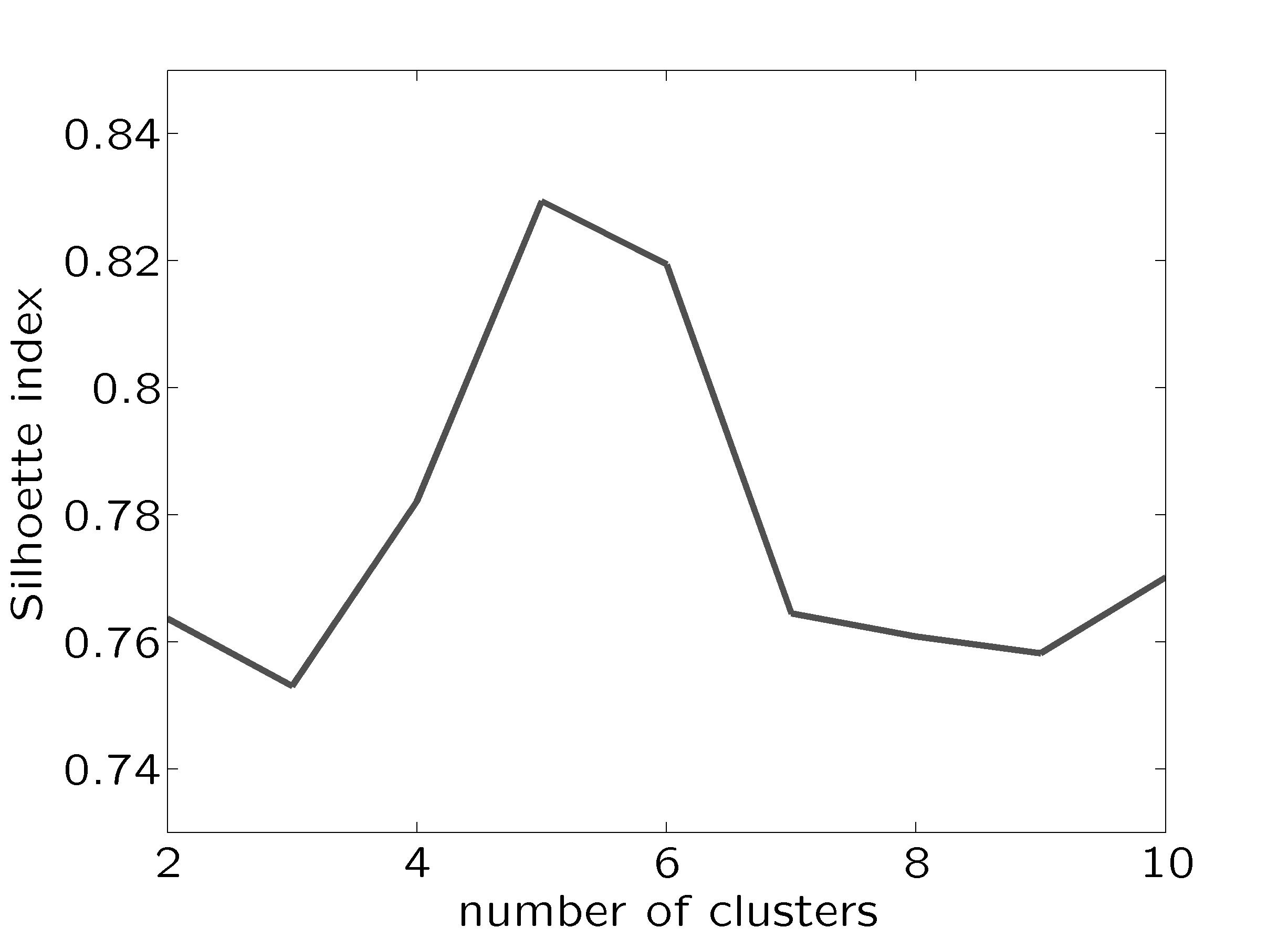}\label{pic:silhouette}}\hspace{0.1in}
  \subfloat[Silhouette measure approximation.]{\includegraphics[width=0.4\textwidth]{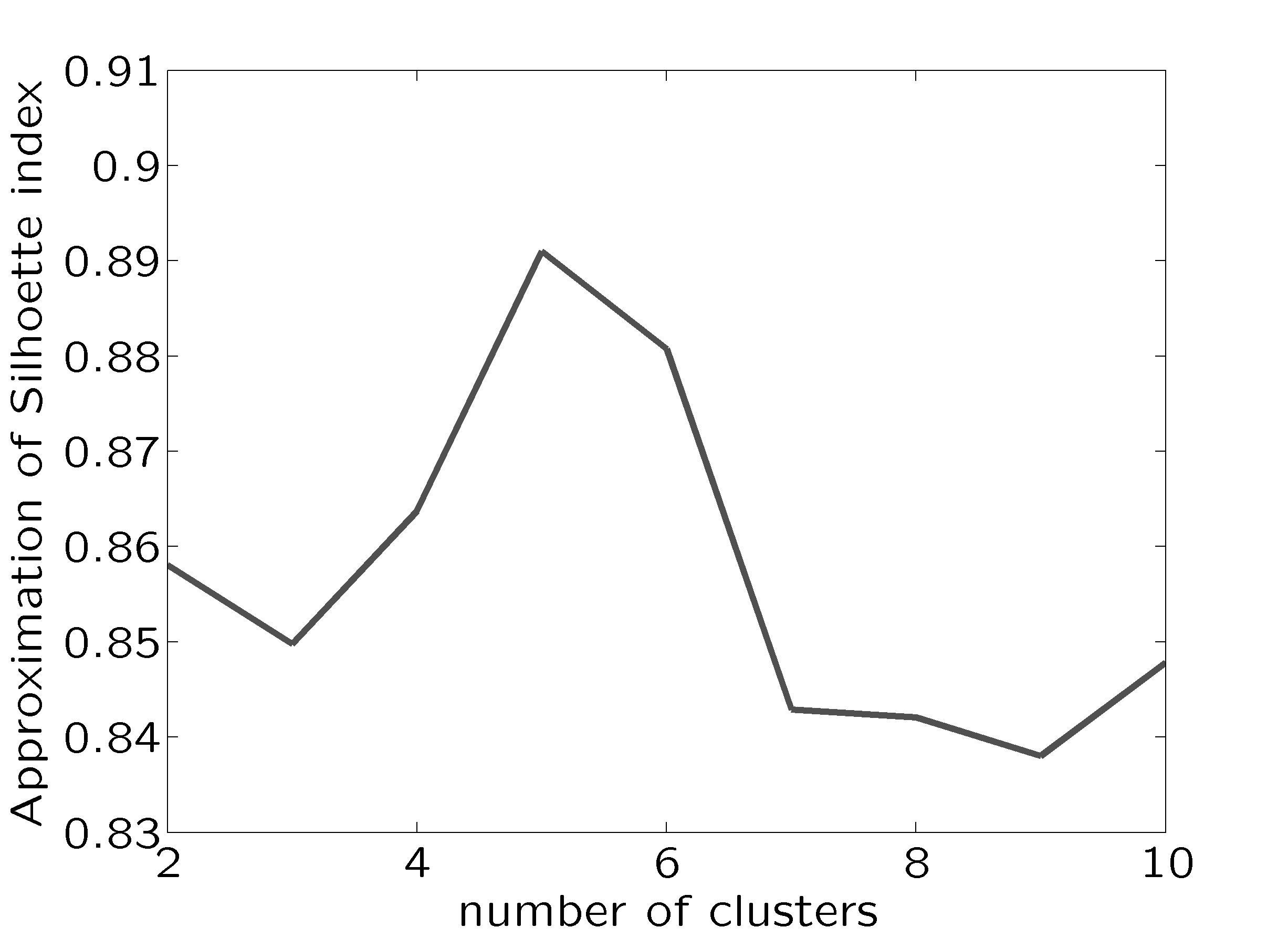}\label{pic:approxisilhouette}}\\
  \caption{Comparison between the Silhouette measure and its approximation. The peaks in the measures correspond to the maximum ratio of inter-cluster similarity to between-cluster diversity. We observe that the peaks of both measures occur at the same number of clusters.}
  \label{pic:silhouettemeasures}
\end{figure} 

Clustering quality measures, such as the Davies–Bouldin index \cite{davies:79} and the Dunn index \cite{dunn:73}, tend to monotonically increase or 
monotonically decrease with the number of clusters. 
This means that, according to these measures, the quality of the clustering increases as the number of clusters grows. 
This fact reflects the monotonic increase in the reconstruction quality, but does not reflect our intuition about the limited number of distinct pencils present in a drawing. 
We decided to use the Silhouette clustering measure \cite{rousseeuw:87}, because it has a local maximum that closely matches our perceived optimal number of colors. 
The idea behind the measure is to optimize the ratio between intra-cluster similarities and between-cluster dissimilarities. 
For each data point $j=1,...,f$, we compute the following two properties: $a(j)$ -- intra-cluster dissimilarity, the average distance from a data point to all 
other data points in the same cluster; and $b(j)$ -- between-cluster dissimilarity, the average distance from a data point to all data points of the closest 
cluster it does not belong to. The Silhouette measure for each data point $j$ is calculated as
\begin{equation*}
  s(j) = \frac{b(j)-a(j)}{\max\{a(j),b(j)\}}, 
\end{equation*}
and the final clustering measure is computed by averaging the above value over all data points. 
The Silhouette measure encourages both cluster separation and cluster uniformity. 
Figure \ref{pic:silhouette} shows the resulting Silhouette measure for the clustering of the drawing of Figure \ref{pic:approximation}. 
We observe a maximum Silhouette measure at five clusters, which corresponds to the point we described before as the most informative split. 
Also, five clusters correspond to a good drawing approximation by color centroids.
 
 
The main disadvantage of this measure is its computational complexity. 
In order to compute $a(j)$ and $b(j)$, we need to compute a dense distance matrix of size $f \times f$ for all pixels in the colored foreground. 
Moreover, we need to run it for all possible hypothetical number of clusters, which in our case ranges from $2$ to $10$ clusters. 
To avoid these computations, we employ an approximation of the Silhouette measure based on information from the cluster centroids, 
and we use this approximation for the remainder of our analysis. We approximate $a(j)$ by the distance from the data point $j$ to its centroid and $b(j)$ 
by the distance to the next closest centroid, keeping the equation for $s(j)$ the same. The approximation of the Silhouette measure for the same drawing is 
shown in Figure \ref{pic:approxisilhouette}. As we see, this computationally inexpensive version of the measure follows the shape of the original measure. 
The absolute values are different, but we are interested in the local maxima that agree in both measures. If we denote by $k$ the number of clusters, 
$k=2,...,10$, now we only need to compute a distance matrix of size $f \times k$ for each clustering with each value of $k$, which is much smaller 
than the original distance matrix. If we denote by $K$ the optimal number of clusters determined by the maximum of the Silhouette measure, the final output 
of the second step consists of $K$ matrices $\mathbf{M}_2^1, \mathbf{M}_2^2,...,\mathbf{M}_2^K$ that group pixels of the same color and $K$ matrices 
$\mathbf{C}_2^1, \mathbf{C}_2^2,...,\mathbf{C}_2^K$ that group the corresponding pixel coordinates.

\subsection{Palette Analysis}
\begin{figure}
  \centering
  \subfloat[CH]{\includegraphics[height=0.05\columnwidth]{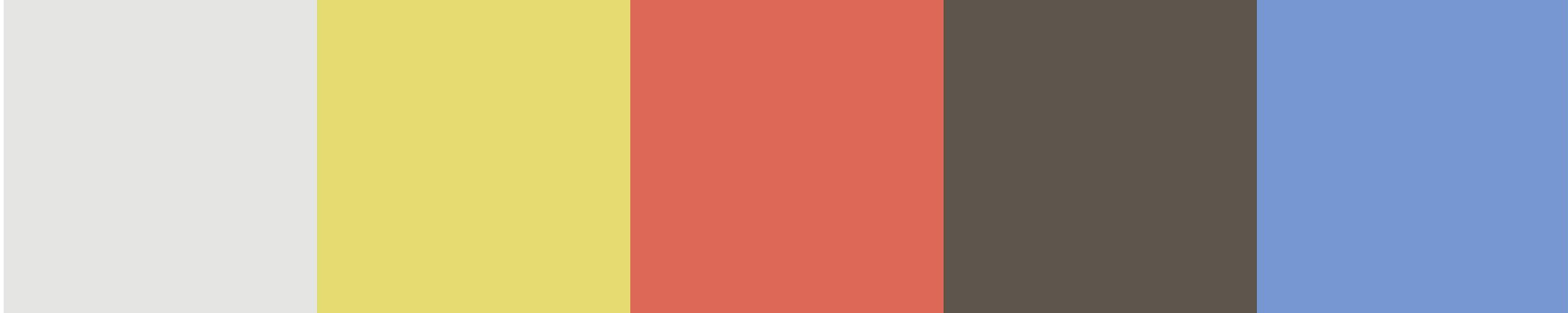}} \\
  \subfloat[JP]{\includegraphics[height=0.05\columnwidth]{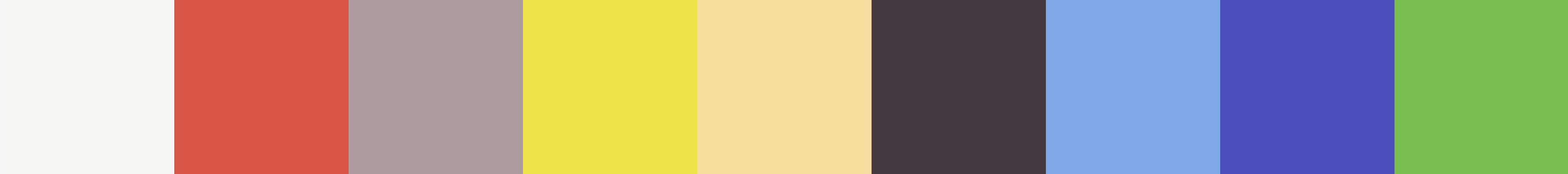}} \\
  \subfloat[RO]{\includegraphics[height=0.05\columnwidth]{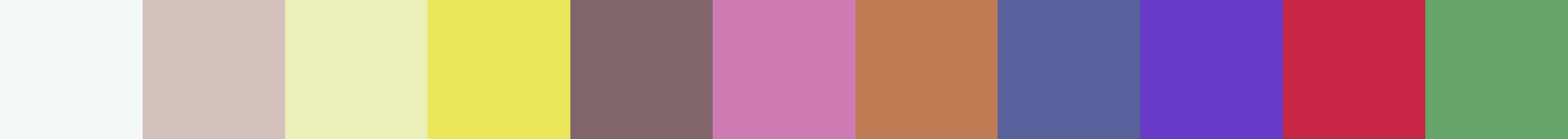}} \\
  \subfloat[RU-bo]{\includegraphics[height=0.05\columnwidth]{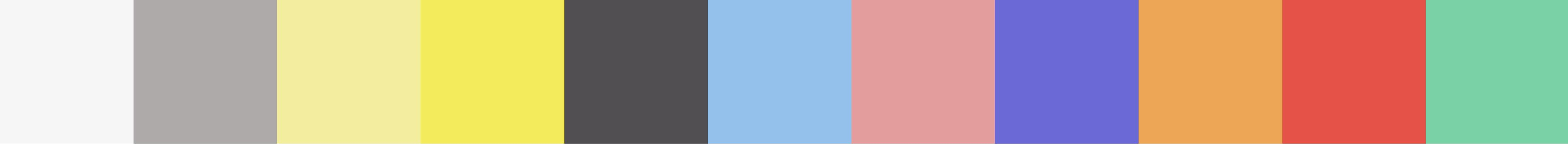}} \\
  \subfloat[RU-sp]{\includegraphics[height=0.05\columnwidth]{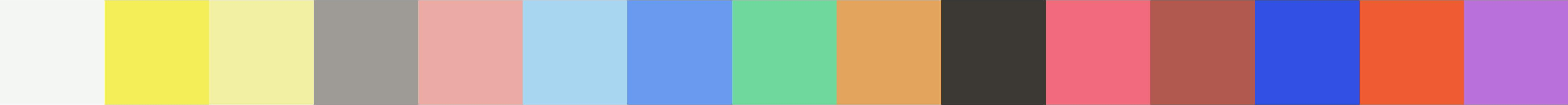}} \\
  \subfloat[US]{\includegraphics[height=0.05\columnwidth]{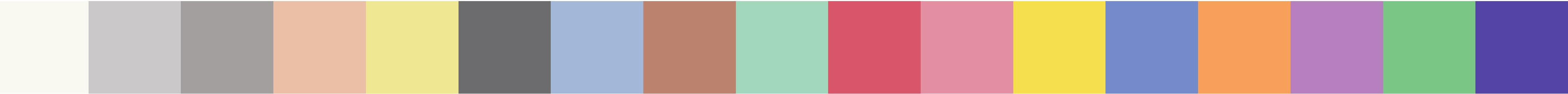}}
  \caption{Extracted color palettes from different countries.} 
  \label{pic:palettes}
\end{figure}

In this section, we analyze the colors from drawings of children of different origins. This gives us the opportunity to analyze the cultural meaning 
of using different colors inside a drawing of God(s). As previously, we group drawings by regions and then we group together all the pencil colors extracted from these images. 
The result is illustrated in Figure \ref{pic:palettes}. The colors are sorted according to their popularity, starting from the most popular color on the left.

We observe that in Switzerland children use only basic colors, such as yellow, red, blue and black. 
At the other extreme, multiple sophisticated color variations are used in the USA. That may be due to the different data collection process. 
Clearly, in most countries the color yellow dominates, as discussed in Section \ref{sec:intensitycolor}. 
Another observation is related to the frequent use of the color gray. The reason is that many children put more emphasis on the drawing procedure, and 
are not interested in coloring their drawing. In these types of drawings the color gray dominates, because these drawings are closer to a gray-scale sketch. 
Blue is moderately popular. 

\subsection{Complexity of drawing}
\begin{figure}
  \centering
  \subfloat[Harris corners.]{\includegraphics[width=0.4\textwidth]{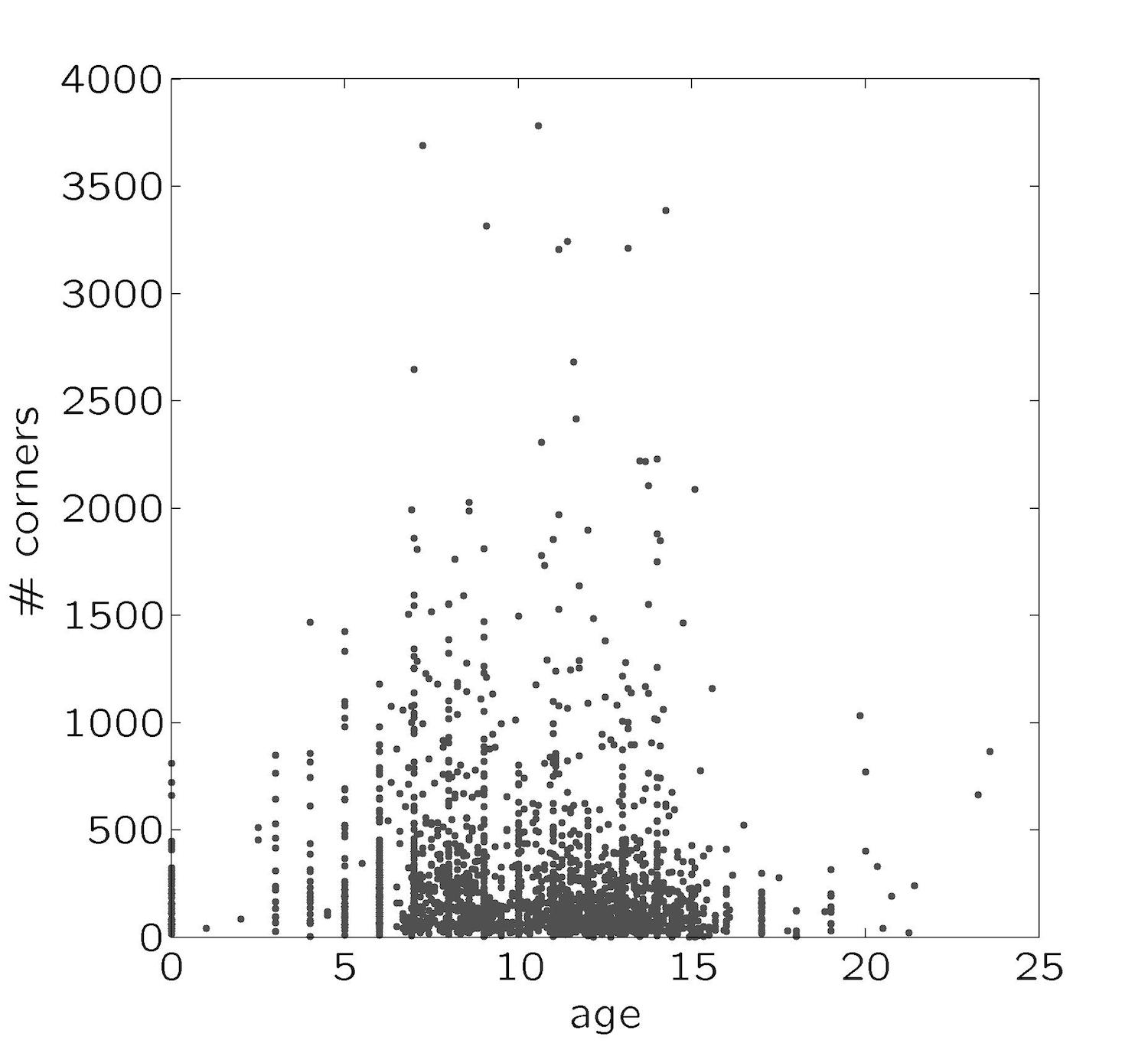}\label{pic:corners}}
  \subfloat[Palette variability.]{\includegraphics[width=0.4\textwidth]{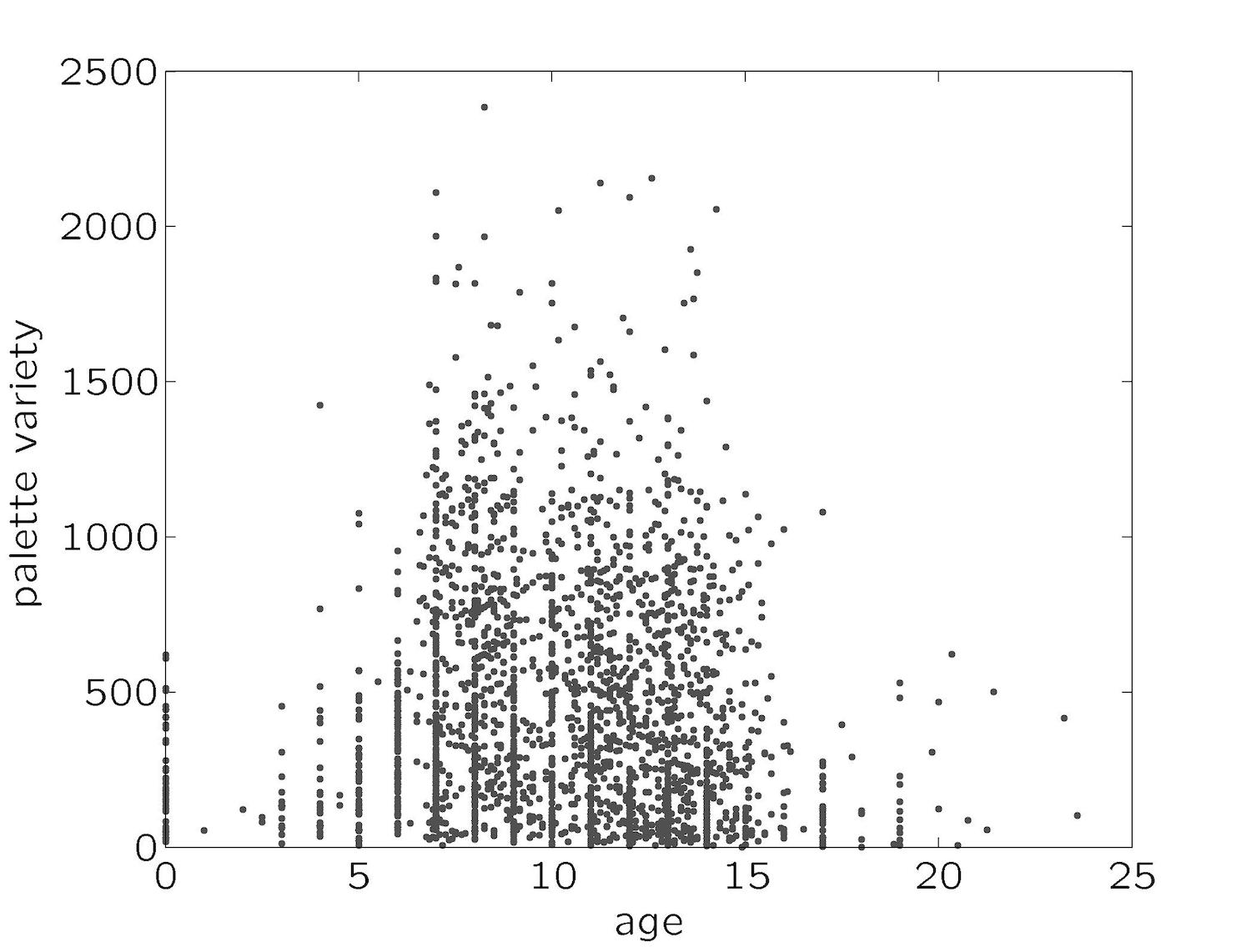}\label{pic:variety}}\\
  \caption{Complexity of childrens' drawings as a function of age using two measures: the number of corners and the palette variability, respectively. We observe a common behavior in both plots.}
  \label{pic:silhouettemeasures}
\end{figure}

Finally, we investigate how the complexity of the drawing depends on the age of the child. We believe and observe empirically that young children draw simple scenes 
(see also \cite{golomb:87}). As they become older, they add more and more details, however, after having reached a certain age they prefer simple scenes for 
the representation of God(s) \cite{harms:44}, \cite{hanisch:96}. 
To find evidence of this hypothesis and to prove it objectively, we visualize in Figure \ref{pic:corners} the distribution of the number of corners 
(according to Harris corner detector) and in Figure \ref{pic:variety} the variability of the palette (the average variance in RGB channels of palette centroids).

 
The number of corners and the variance in the palette can serve as indirect evidence of the complexity of the scene. 
Figures \ref{pic:corners} and \ref{pic:variety} show the distribution of these measures for all the drawings. Each drawing is represented as a point in the figures.
Both of the measures are consistent with the above hypothesis on the drawing complexity in varying age groups.

\section{Discussion and Conclusions} 
\label{sec:concl}
In this article, we have introduced a set of tools for automatic pattern analysis of children's drawings of God(s) and other supernatural agents. 
To our knowledge, such a study has never been conducted before on a dataset similar to ours. We have shown, using techniques from computer vision and scientific imaging, 
how to detect developmental and cross-cultural patterns for the needs of researchers in psychology of religion. 

We find that children have the tendency to draw above the center of the page. This gives an indication that they believe God(s) does not live among us, but 
above us towards the sky. We statistically confirm that the split of children's age into three groups ($(1,7]$, $(7,11]$, $>11$) \cite{piaget:00} gives the most 
reliable information about the changes in their behaviour. Another important finding is the fact that the formulation of the task given to the children 
plays a very important role and can significantly influence the results of the analysis. 

Concerning color choice, we observe that children use yellow to draw not only natural objects, such as the sun, but also parts of the supernatural 
agent. Green usually corresponds to elements of nature, because it is often distributed at the bottom of the page. This supports the hypotheses in 
psychology of religion about the roles of these colors: green is used for earthly contexts, yellow is used for God(s) and sun representations. 

Children from different cultures prefer different colors. Furthermore, the amount of colors used changes across religions and 
cultural backgrounds. However, there are still universal drawing strategies: gray and yellow are the most popular colors. Gray usually corresponds to grayscale drawings with no color. 
Further investigation and more data are needed to produce more concrete conclusions about the importance of the different colors among cultures. 

Finally, we observe that children tend to draw more complex scenes as they grow older. However, after some age (approximately $13$ years old), they start 
drawing simple scenes again. The investigation of this behavior is one of our goals, and a more thorough investigation of the actual 
objects and their complexity will shed more light into the behavior and evolution of children's drawings through age. This is an important direction for future research. 

The research project on children's drawings of God(s) is ongoing. The continuously growing dataset will enable researchers to perform on-line 
data mining analysis, similar to the one presented in this work. We believe that the public availability of this dataset will help promote this direction of research.

\bibliographystyle{ACM-Reference-Format-Journals}
\bibliography{litterature}


\begin{thebibliography}{00}


\ifx \showCODEN    \undefined \def \showCODEN     #1{\unskip}     \fi
\ifx \showDOI      \undefined \def \showDOI       #1{{\tt DOI:}\penalty0{#1}\ }
  \fi
\ifx \showISBNx    \undefined \def \showISBNx     #1{\unskip}     \fi
\ifx \showISBNxiii \undefined \def \showISBNxiii  #1{\unskip}     \fi
\ifx \showISSN     \undefined \def \showISSN      #1{\unskip}     \fi
\ifx \showLCCN     \undefined \def \showLCCN      #1{\unskip}     \fi
\ifx \shownote     \undefined \def \shownote      #1{#1}          \fi
\ifx \showarticletitle \undefined \def \showarticletitle #1{#1}   \fi
\ifx \showURL      \undefined \def \showURL       #1{#1}          \fi

\bibitem[\protect\citeauthoryear{Brandt}{Brandt}{2010}]%
        {brandt:10}
{Pierre-Yves Brandt}. 2010.
\newblock {\em {Des enfants dessinent Dieu: oiseaux, mangas, soleils et
  couleurs...}}
\newblock Labor et Fides.
\newblock


\bibitem[\protect\citeauthoryear{Brandt, Kagata~Spitteler, and
  Gilli{\`e}ron~Pal{\'e}ologue}{Brandt et~al\mbox{.}}{2009}]%
        {brandt:09}
{Pierre-Yves Brandt}, {Yuko Kagata~Spitteler}, {and} {Christiane
  Gilli{\`e}ron~Pal{\'e}ologue}. 2009.
\newblock \showarticletitle{{La repr{\'e}sentation de Dieu: Comment des enfants
  japonais dessinent Dieu}}.
\newblock {\em Archives de Psychologie 74\/} (2009), 171--203.
\newblock


\bibitem[\protect\citeauthoryear{Comaniciu and Meer}{Comaniciu and
  Meer}{2002}]%
        {comaniciu:02}
{Dorin Comaniciu} {and} {Peter Meer}. 2002.
\newblock \showarticletitle{{Mean shift: A robust approach toward feature space
  analysis}}.
\newblock {\em Pattern Analysis and Machine Intelligence, IEEE Transactions
  on\/} {24}, 5 (2002), 603--619.
\newblock


\bibitem[\protect\citeauthoryear{Dandarova}{Dandarova}{2013}]%
        {dandarova:13}
{Zhargalma Dandarova}. 2013.
\newblock \showarticletitle{{Le dieu des enfants: entre l'universel et le
  contextuel}}.
\newblock In {\em Psychologie du d{\'e}veloppement religieux. Questions
  classiques et perspectives contemporaines}, {Pierre-Yves Brandt} {and} {James
  M.~Day} (Eds.). Labor et Fides, 159--187.
\newblock


\bibitem[\protect\citeauthoryear{Davies and Bouldin}{Davies and
  Bouldin}{1979}]%
        {davies:79}
{David~L. Davies} {and} {Donald~W. Bouldin}. 1979.
\newblock \showarticletitle{{A cluster separation measure}}.
\newblock {\em Pattern Analysis and Machine Intelligence, IEEE Transactions
  on\/} 2 (1979), 224--227.
\newblock


\bibitem[\protect\citeauthoryear{Dempster, Laird, and Rubin}{Dempster
  et~al\mbox{.}}{1977}]%
        {dempster:77}
{Arthur~P. Dempster}, {Nan~M. Laird}, {and} {Donald~B. Rubin}. 1977.
\newblock \showarticletitle{{Maximum likelihood from incomplete data via the EM
  algorithm}}.
\newblock {\em Journal of the Royal statistical Society\/} {39}, 1 (1977),
  1--38.
\newblock


\bibitem[\protect\citeauthoryear{Dunn}{Dunn}{1973}]%
        {dunn:73}
{Joseph~C. Dunn}. 1973.
\newblock \showarticletitle{{A fuzzy relative of the ISODATA process and its
  use in detecting compact well-separated clusters}}.
\newblock {\em Journal of Cybernetics\/} {3}, 3 (1973), 32--57.
\newblock


\bibitem[\protect\citeauthoryear{Eshleman, Dickie, Merasco, Shepard, and
  Johnson}{Eshleman et~al\mbox{.}}{1999}]%
        {eshleman:99}
{Amy~K. Eshleman}, {Jane~R. Dickie}, {Dawn~M. Merasco}, {Amy Shepard}, {and}
  {Melissa Johnson}. 1999.
\newblock \showarticletitle{{Mother God, father God: Children's perceptions of
  God's distance}}.
\newblock {\em The International Journal for the Psychology of Religion\/} {9},
  2 (1999), 139--146.
\newblock


\bibitem[\protect\citeauthoryear{Golomb}{Golomb}{1987}]%
        {golomb:87}
{Claire Golomb}. 1987.
\newblock \showarticletitle{{The Development of Compositional Strategies in
  Children's Drawings}}.
\newblock {\em Visual Arts Research\/} {13}, 2 (1987), 42--52.
\newblock


\bibitem[\protect\citeauthoryear{Goodman and Manierre}{Goodman and
  Manierre}{2008}]%
        {goodman:08}
{Geoff Goodman} {and} {Amy Manierre}. 2008.
\newblock \showarticletitle{{Representations of God uncovered in a spirituality
  group of borderline inpatients}}.
\newblock {\em International Journal of Group Psychotherapy\/} {58}, 1 (2008),
  1--15.
\newblock


\bibitem[\protect\citeauthoryear{Hanisch}{Hanisch}{1996}]%
        {hanisch:96}
{Helmut Hanisch}. 1996.
\newblock {\em {Die zeichnerische Entwicklung des Gottesbildes bei Kindern und
  Jugendlichen: eine empirische Vergleichsuntersuchung mit religi{\"o}s und
  nicht-religi{\"o}s Erzogenen im Alter von 7-16 Jahren}}.
\newblock Calwer Verlag.
\newblock


\bibitem[\protect\citeauthoryear{Hargreaves, Jones, and Martin}{Hargreaves
  et~al\mbox{.}}{1981}]%
        {hargreaves:81}
{David~J. Hargreaves}, {Philip~M. Jones}, {and} {Diane Martin}. 1981.
\newblock \showarticletitle{{The air gap phenomenon in children's landscape
  drawings}}.
\newblock {\em Journal of Experimental Child Psychology\/} {32}, 1 (1981),
  11--20.
\newblock


\bibitem[\protect\citeauthoryear{Harms}{Harms}{1944}]%
        {harms:44}
{Ernest Harms}. 1944.
\newblock \showarticletitle{The Development of Religious Experience in
  Children}.
\newblock {\it Amer. J. Sociology} {50}, 2 (1944), 112--122.
\newblock


\bibitem[\protect\citeauthoryear{Henze}{Henze}{1988}]%
        {henze:88}
{Norbert Henze}. 1988.
\newblock \showarticletitle{{A multivariate two-sample test based on the number
  of nearest neighbor type coincidences}}.
\newblock {\em The Annals of Statistics\/} (1988), 772--783.
\newblock


\bibitem[\protect\citeauthoryear{Kay and Ray}{Kay and Ray}{2004}]%
        {kay:04}
{William~K. Kay} {and} {Liz Ray}. 2004.
\newblock \showarticletitle{{Concepts of God: The salience of gender and age}}.
\newblock {\em Journal of Empirical Theology\/} {17}, 2 (2004), 238--251.
\newblock


\bibitem[\protect\citeauthoryear{K{\"o}se}{K{\"o}se}{2008}]%
        {kose:08}
{Sacit K{\"o}se}. 2008.
\newblock \showarticletitle{{Diagnosing student misconceptions: using drawings
  as a research method}}.
\newblock {\em World Applied Sciences Journal\/} {3}, 2 (2008), 283--293.
\newblock


\bibitem[\protect\citeauthoryear{Ladd, McIntosh, and Spilka}{Ladd
  et~al\mbox{.}}{1998}]%
        {ladd:98}
{Kevin~L. Ladd}, {Daniel~N. McIntosh}, {and} {Bernard Spilka}. 1998.
\newblock \showarticletitle{{Children's God concepts: Influences of
  denomination, age, and gender}}.
\newblock {\em The International Journal for the Psychology of Religion\/} {8},
  1 (1998), 49--56.
\newblock


\bibitem[\protect\citeauthoryear{Littleton and Wood}{Littleton and
  Wood}{2006}]%
        {wood:06}
{Karen Littleton} {and} {Clare Wood}. 2006.
\newblock \showarticletitle{{Psychology and education: Understanding teaching
  and learning}}.
\newblock In {\em Developmental psychology in action}, {Clare Wood}, {Karen
  Littleton}, {and} {Kieron Sheehy} (Eds.). Wiley-Blackwell, 193--229.
\newblock


\bibitem[\protect\citeauthoryear{Mach{\'o}n}{Mach{\'o}n}{2013}]%
        {machon:13}
{Antonio Mach{\'o}n}. 2013.
\newblock {\em {CHILDREN'S DRAWINGS: The Genesis and Nature of Graphic
  Representation. A Developmetnal Study}}.
\newblock Editorial Fibulas.
\newblock


\bibitem[\protect\citeauthoryear{MacQueen}{MacQueen}{1967}]%
        {macqueen:67}
{James MacQueen}. 1967.
\newblock \showarticletitle{{Some methods for classification and analysis of
  multivariate observations}}. In {\em Proceedings of the fifth Berkeley
  symposium on mathematical statistics and probability}, Vol.~1. 281--297.
\newblock


\bibitem[\protect\citeauthoryear{Newberg and Waldman}{Newberg and
  Waldman}{2010}]%
        {newberg:10}
{Andrew Newberg} {and} {Mark~Robert Waldman}. 2010.
\newblock {\em {How God changes your brain: Breakthrough findings from a
  leading neuroscientist}}.
\newblock Random House LLC.
\newblock


\bibitem[\protect\citeauthoryear{{\'O}skarsd{\'o}ttir, Stougaard, Fleischer,
  Jeronen, L{\"u}tzen, and Kr{\aa}kenes}{{\'O}skarsd{\'o}ttir
  et~al\mbox{.}}{2011}]%
        {oskarsdottir:11}
{Gunnhildur {\'O}skarsd{\'o}ttir}, {Birgitte Stougaard}, {Ane Fleischer}, {Eila
  Jeronen}, {Finnur L{\"u}tzen}, {and} {Roar Kr{\aa}kenes}. 2011.
\newblock \showarticletitle{{Children's ideas about the human body - A Nordic
  case study}}.
\newblock {\em NorDiNa - Nordic Studies in Science Education\/} {7}, 2 (2011),
  179--188.
\newblock


\bibitem[\protect\citeauthoryear{Piaget and Inhelder}{Piaget and
  Inhelder}{2000}]%
        {piaget:00}
{Jean Piaget} {and} {B{\"a}rbel Inhelder}. 2000.
\newblock {\em The Psychology Of The Child}.
\newblock NY: Basic Books.
\newblock


\bibitem[\protect\citeauthoryear{Pitts}{Pitts}{1977}]%
        {pitts:77}
{Peter Pitts}. 1977.
\newblock \showarticletitle{{Drawing pictures of God}}.
\newblock {\em Learning for Living\/} {16}, 3 (1977), 123--129.
\newblock


\bibitem[\protect\citeauthoryear{Reiss and Tunnicliffe}{Reiss and
  Tunnicliffe}{2001}]%
        {reiss:01}
{Michael~J. Reiss} {and} {Sue~Dale Tunnicliffe}. 2001.
\newblock \showarticletitle{{Students' understandings of human organs and organ
  systems}}.
\newblock {\em Research in Science Education\/} {31}, 3 (2001), 383--399.
\newblock


\bibitem[\protect\citeauthoryear{Rizzuto}{Rizzuto}{1981}]%
        {rizzuto:81}
{Ana-Marie Rizzuto}. 1981.
\newblock {\em {The birth of the living God: A psychoanalytic study}}.
\newblock University of Chicago Press.
\newblock


\bibitem[\protect\citeauthoryear{Rousseeuw}{Rousseeuw}{1987}]%
        {rousseeuw:87}
{Peter~J. Rousseeuw}. 1987.
\newblock \showarticletitle{{Silhouettes: a graphical aid to the interpretation
  and validation of cluster analysis}}.
\newblock {\it J. Comput. Appl. Math.}  {20} (1987), 53--65.
\newblock


\bibitem[\protect\citeauthoryear{Shi and Malik}{Shi and Malik}{2000}]%
        {shi:00}
{Jianbo Shi} {and} {Jitendra Malik}. 2000.
\newblock \showarticletitle{{Normalized cuts and image segmentation}}.
\newblock {\em Pattern Analysis and Machine Intelligence, IEEE Transactions
  on\/} {22}, 8 (2000), 888--905.
\newblock


\bibitem[\protect\citeauthoryear{Stork}{Stork}{2009}]%
        {stork:09}
{David~G Stork}. 2009.
\newblock \showarticletitle{{Computer vision and computer graphics analysis of
  paintings and drawings: An introduction to the literature}}. In {\em Computer
  Analysis of Images and Patterns}. 9--24.
\newblock
\showURL{%
\url{http://link.springer.com/chapter/10.1007/978-3-642-03767-2_2}}


\bibitem[\protect\citeauthoryear{Strommen}{Strommen}{1995}]%
        {strommen:95}
{Erik Strommen}. 1995.
\newblock \showarticletitle{{Lions and tigers and bears, oh my! Children's
  conceptions of forests and their inhabitants}}.
\newblock {\em Journal of Research in Science Teaching\/} {32}, 7 (1995),
  683--698.
\newblock


\bibitem[\protect\citeauthoryear{Tamm}{Tamm}{1996}]%
        {tamm:96}
{Maare~E. Tamm}. 1996.
\newblock \showarticletitle{{The meaning of God for children and adolescents -
  a phenomenographic study of drawings}}.
\newblock {\em British Journal of Religious Education\/} {19}, 1 (1996),
  33--44.
\newblock


\bibitem[\protect\citeauthoryear{Winner}{Winner}{2006}]%
        {winner:06}
{Ellen Winner}. 2006.
\newblock \showarticletitle{{Development in the arts: Drawing and music}}.
\newblock In {\em Handbook of Child Psychology}, {Deanna Kuhn} {and} {Robert
  Siegler} (Eds.). Wiley, 859--904.
\newblock


\end{thebibliography}

\end{document}